# Three-dimensional visualization of X-ray Micro-CT with large-scale datasets: Efficiency and Accuracy for Real-time Interaction

Yipeng Yin, Rao Yao*, Qingying Li, Dazhong Wang, Hong Zhou, Zhijun Fang, Jianing Chen, Longjie Qian, Mingyue Wu

Aviation College, Shanghai University of Engineering Science, Shanghai 201620, China

Corresponding E-mail: yr_sues@163.com

**ABSTRACT** - As Micro-CT technology continues to refine its characterization of material microstructures, industrial CT ultra-precision inspection is generating increasingly large datasets, necessitating solutions to the trade-off between accuracy and efficiency in the 3D characterization of defects during ultra-precise detection. This article provides a unique perspective on recent advances in accurate and efficient 3D visualization using Micro-CT, tracing its evolution from medical imaging to industrial non-destructive testing (NDT). Among the numerous CT reconstruction and volume rendering methods, this article selectively reviews and analyzes approaches that balance accuracy and efficiency, offering a comprehensive analysis to help researchers quickly grasp highly efficient and accurate 3D reconstruction methods for microscopic features. By comparing the principles of computed tomography and advancements in microstructural technology, the progress of CT reconstruction algorithms from analytical methods to deep learning techniques, and improvements in volume rendering concerning algorithms, acceleration, and data reduction, this article also explores advanced lighting models for high-accuracy, photorealistic, and efficient volume rendering. Furthermore, this article envisions potential directions in CT reconstruction and volume rendering. It aims to guide future research in quickly selecting efficient and precise methods and developing new ideas and approaches for real-time online monitoring of internal material defects through virtual-physical interaction, for applying digital twin model to structural health monitoring (SHM).

## 1. INTRODUCTION

3D Reconstruction is one of the important research directions in the field of visualization. With the theory of graphic semiotics and computer image processing technology, three-dimensional reconstruction has also developed rapidly, using computer graphics to help people understand large and complex scientific concepts or results through the creation of visual images.

In the fields of industrial flaw detection and medicine, three-dimensional reconstruction is combined with X-ray, ultrasound, nuclear magnetic resonance and other detection methods, and three-dimensional reconstruction represented by volume rendering provides researchers with a more efficient and easier to understand research and analysis tools[1, 2].Volume rendering refers to a series of techniques that display a three-dimensional dataset as a two-dimensional image. By classifying the spatial information of the volume data and projecting it onto the screen, researchers can better understand the different characteristics of the volume data. These technologies fall into two main categories: Indirect volume rendering and Direct Volume Rendering (DVR)[3]. Indirect volume rendering is a volume-based representation of polygon surfaces, usually using equipotential surfaces of volume data to generate triangular surfaces, which can only simply reconstruct the object surface, however cannot be reconstructed for some details. DVR is a direct mapping of the volume equation and is implemented in the form of volume ray projection. Since it does not depend on the intermediate geometric polygon elements, it is highly efficient. DVR has become a vital and important technology in medical image visualization, scientific computing simulation and other research fields, and has broad application prospects. Volume rendering, as a post-processing technology, often needs to be combined with volume data acquisition, CT [4] is a mainstream medical image tool and an industrial non-destructive testing method [5, 6], which can provide the internal information of the detected object, and volume rendering can make more efficient use of CT volume data and provide visual results.

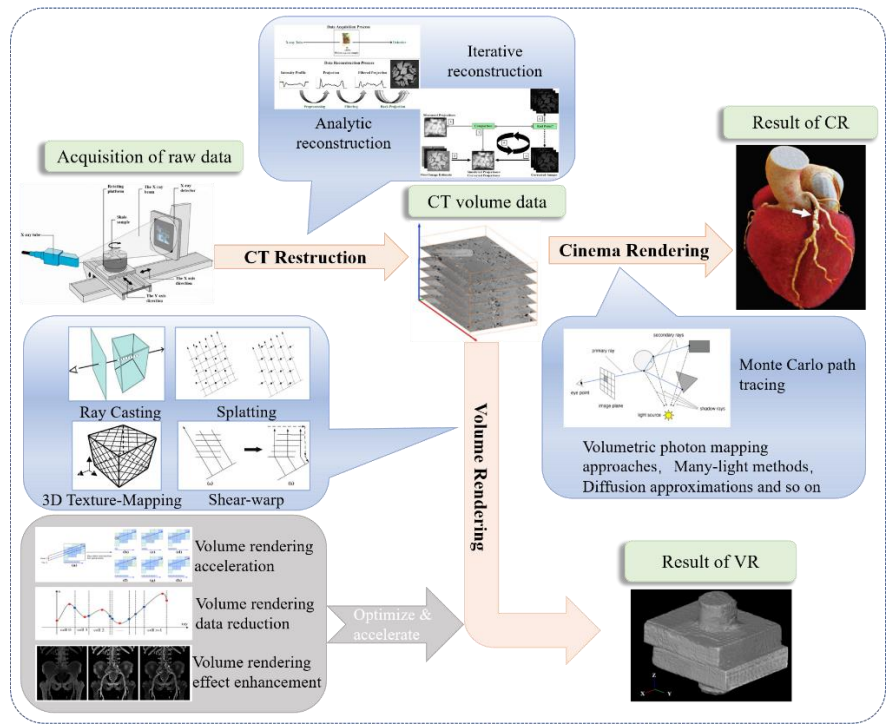

Figure 1: Main content of the review article.

Micro-CT is a kind of high resolution microscopic non-destructive imaging technology [7]. The amount of data obtained is often very large, rendering quality and rendering efficiency have become the two core issues that people are concerned about to accurately display the volume data information and to meet the speed requirements of interactive visual analysis. With the improvement of the precision of Micro-CT required to observe microscopic damage, the amount of data obtained is becoming larger and larger. Theoretically, the continuous encryption of sampling points can accurately represent each physical feature in volume rendering. However, both sampling calculation and image synthesis in volume rendering are very time-consuming, and the calculation cost of both is proportional to the total number of sampling points, so increasing the sampling points will inevitably cause huge calculation cost and lead to too long rendering time. However as with all other post-processing techniques, such as MIP or SSD, visualizing imaging data using VR does not add information to the reader that was present in the original source image. It is always a choice for different representations of imaging data, so the accuracy of the volume data set obtained by CT and the reconstruction speed are also important factors affecting the volume rendering effect. Aiming to adapt to the amount of increasing data and the increasingly complex physical characteristics in the data, researchers have conducted in-depth research on CT reconstruction algorithm and volume rendering algorithm from many aspects to continuously improve reconstruction efficiency, rendering efficiency and enhance rendering effect. Furthermore, with enhancing hardware computational power, a new volume rendering algorithm using global illumination models and path tracing, distinct from traditional local illumination models, is becoming a hot topic in current research, known as CR (cinema rendering). Compared to traditional algorithms, CR offers better realism and provides more accurate depth information, and it is rapidly developing in application fields from medical to industrial.

Micro-CT equipment's CT reconstruction algorithms and volume rendering algorithms involve completely different underlying disciplines and algorithms. The CT reconstruction algorithm serves as a prerequisite for the volume rendering algorithm in the three-dimensional imaging process. Only by providing a sufficiently excellent volume dataset can the volume rendering algorithm achieve better rendering effects. Therefore, to achieve better visualization effects, researchers need to optimize these two algorithms separately and organically combine them to obtain the best results. At the same time, as the hardware foundation for three-dimensional imaging, CT equipment also determines the upper limit of imaging quality, and the minimum resolution of the detector also represents the minimum resolution that CT images can achieve.

Within the industry, a systematic review of the three-dimensional reconstruction technologies for microscopic CT and their classifications pertaining to big CT datasets has yet to be conducted. This article identifies and integrates the nexus between CT reconstruction and volume rendering. Anchored in the cutting-edge advancements of microscopic CT and medical CT three-dimensional reconstructions, this article provides a comprehensive review of existing technologies, introduces original insights, and projects the future development of efficient and high-precision algorithms for three-dimensional reconstruction from medical to industry. The review is divided into four parts as in Figure 1: the first part introduces the most advanced Micro-CT equipment and its principles, pointing out that the evolution of hardware determines the quality, volume, difficulty, efficiency and accuracy of image reconstruction; the second part discusses the development and current status of CT reconstruction algorithms, focusing on solving the problems of efficiency and accuracy; the third part focuses on the acceleration and optimization of traditional volume rendering algorithms; and the fourth part reviews the current state of research on volume rendering algorithms that use global illumination models and path tracing which has significant improvements in the precision, realism and efficiency.

## 2. PRINCIPLES OF COMPUTED TOMOGRAPHY AND EVOLUTION OF MICROSTRUCTURAL TECHNIQUES COMPARISON

The mathematical idea of Computed Tomography (CT) was first discovered by J. Radon, who proposed a mathematical method of graph reconstruction in his doctoral thesis which is mapping a given function to a Radon transform of a line integral and reconstructing the line integral into an inverse Radon transform of an image [8]. With the development of computer science in the 1960s, Cormack discovered the possibility of applying the Radon transform to radiology, and in 1963 proposed the application of the Randon transform to medical image reconstruction, namely CT, using reconstruction formulas to obtain absorption coefficients [9].

Table 1: Advantage and disadvantage of different microscopic technique

| Microscopic technique | Advantage | Disadvantage | Reference |
|---|---|---|---|
| Mercury Intrusion Porosity Determination (MIP) | High measurement accuracy; Easy to use. | May destroy sample; Only for porous materials. | [10, 14] |
| X-ray diffraction (XRD) | Qualitative analysis of sample composition; Faster process. | Sample Requirements; Crystal Structure; Unable to detect low-level impurities. | [11, 15, 16] |
| Scanning electron microscopy (SEM) | High resolution; Large depth of field and high magnification. | Unable to provide sample internal information; Lack of height orientation information. | [12, 17, 18] |
| Nuclear magnetic resonance (NMR) | Non-destructive; Quantitative analysis of sample composition. | Unable to provide microstructure; Low sensitivity. | [13, 19] |
| Micro tomography (Micro CT) | Provide 3-D image of sample; Non-invasive. | Requires high computer computing power; Sample size is limited. | [20, 21] |

With the development of science and technology and the exploration of microstructure characteristics of materials and components, mercury intrusion porosity determination (MIP), X-ray diffraction (XRD), scanning electron microscopy (SEM), transmission electron microscopy (TEM), nuclear magnetic resonance (NMR) and many other methods have been proposed. MIP is one of the most widely used techniques for determining the pore size distribution from capillary to air gap of a sample, which is easy to implement [10]. XRD can be used to examine the compounds, minerals, and clays present in powdered samples to qualitatively determine the substance [11]. SEM can also provide information on the microstructure of the sample surface by using high-resolution images [12]. NMR can be used to examine extremely small components, such as gels and capillary holes [13]. However, despite its advantages, there are still limitations to the effectiveness of the above methods in identifying microstructure. MIP can only measure open-hole structures and measure damaged specimens during pressure loading. XRD is only reliable when studying crystalline compounds. SEM and other optical microscopes (such as TEM) usually only provide information of the two-dimensional surface images, while NMR can only provide quantitative information on the components, which is difficult to study the spatial characteristics of the microstructure. The specific comparisons are presented in Table 1.

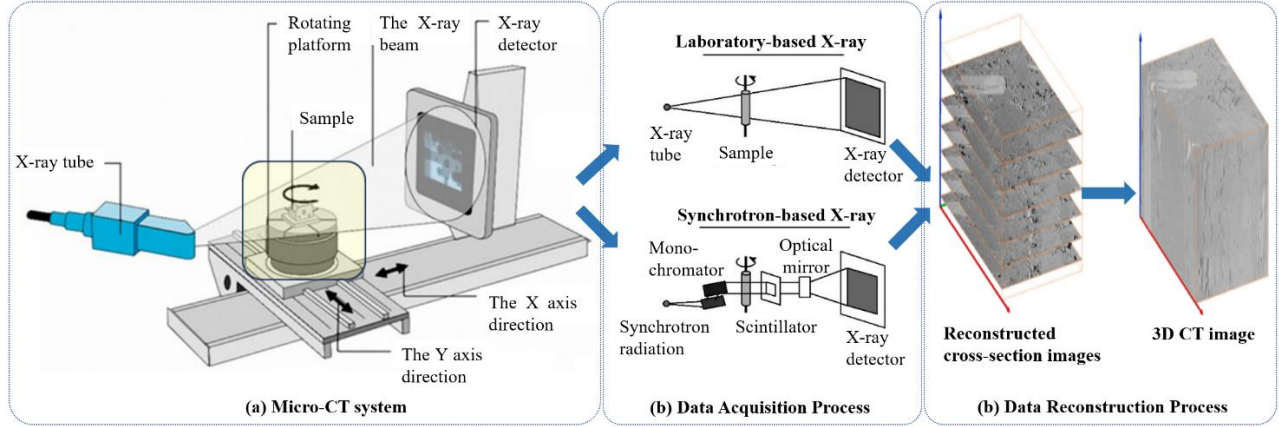

Figure 2: Schematic diagram of Micro-CT system and its 3D reconstruction.

To solve the above limitations, X-ray micro-computed tomography (Micro-CT) has become the first choice to solve such problems. Micro-CT is a radiographic imaging technique that can identify the morphological structure of the target material without damaging the specimen，which can suit to the characteristics of volume rendering that can display internal information of objects better [20]. Micro-CT can be used to generate a series of cross-sectional (2D) images of the material without damaging the material. Typically, a set of reconstructed images represented by pixels (8-bit or 16-bit) can be achieved with Micro-CT. Micro-CT image processing converts 8-bit or 16-bit images into binary or segmented images to classify specific compounds of the specimen, and the resulting segmented images are used to examine the properties of compounds in the target specimen. The 3D microstructure model is generated by stacking 2D segmentation images, which enables multi-directional, volume and other advanced measurements. The entire process is displayed in Figure 2.

## 2.1 Principles of X-ray Attenuation in Micro-CT Imaging

X-rays have a higher energy than other electromagnetic waves and it can penetrate objects to be detected by the detector. Therefore, X-rays are the main type of radiation used in Micro-CT. X-rays at low energy levels cannot penetrate thicker materials, while X-rays at high energy levels cannot provide low-contrast information. The X-rays used for imaging have a range between 12.4 and 124keV, where the wavelength ranges from 0.1 to 0.01nm. The change in X-ray intensity as it passes through a material is described by the Lambert-Beer law [22] as formula (1):

$$I = I_0 e^{-\int \mu(x)dx} \qquad （1）$$

where $I_0$ is the incident intensity of the X-ray, I is intensity after passing through a material of a certain thickness (x), which is characterized by a linear attenuation coefficient μ(x). Note that in this formula, it is assumed that the X-ray source passing through the object is monochromatic and the medium is a vacuum. It is clear from formula (1) that materials with higher μ values have greater attenuation of X-ray photons, while materials with lower μ values have less attenuation of X-ray photons. In this respect, the input X-ray photon flux will remain constant compared to the output X-ray photon flux in a vacuum, because the μ value of a vacuum is almost zero.

Inhomogeneous objects are common in real-world industrial CT applications, consisting of multiple materials or phases with different attenuation coefficients. The object can be divided into a matrix of material volume pixels (voxels) to calculate the total attenuation characteristics. Each volume pixel (voxel) is considered small enough to approach zero, allowing each element to be treated as a homogeneous object. Beer-Lambert law can be reformulated as formula (2):

$$I = I_0 e^{-\mu_1 \triangle x} e^{-\mu_2 \triangle x} e^{-\mu_3 \triangle x} \cdots e^{-\mu_n \triangle x} = I_0 e^{-\sum_{n=1}^{N} \mu_n \triangle x} \qquad （2）$$

Mathematically, the equation can be further derived as an integral over the length of the object L, as shown in formula (3):

$$p = -\ln\left(\frac{I}{I_0}\right) = \int_L \mu(x)dx \qquad （3）$$

Where p is defined as a projected measurement. It is also the line integral of the attenuation coefficient L along the X-ray path, equal to the ratio of the input X-ray intensity to the output X-ray intensity after a logarithmic operation, sometimes defined as the Radon transform.

## 2.2   Components and Micro-CT Systems Spatial Resolution

A typical Micro-CT system consists of four main parts: a Micro-focus X-ray source, a rotating table, a flat panel detector, and a computer system. The schematic diagram of the Micro-CT system is shown in Fig. 2. Attenuation of X-rays by the Beer-Lambert law of a non-uniform object applied in most common and realistic industrial conditions. The non-uniform object can be subdivided into multiple elements for total attenuation coefficient calculation. Each element demonstrates attenuation in the case of a homogeneous object.

When the X-ray beam passes through the specimen, the X-ray projection can be obtained by measuring the intensity attenuation. Then, two-dimensional X-ray projections of different angular positions are obtained by rotating the sample. When used, X-ray Micro-CT can realize geometric scaling of the collected CT data by moving the rotating table in the X and y axis directions, thus changing the spatial resolution of CT. The spatial resolution of CT imaging is shown as formula (4):

$$R_s = \frac{\sqrt{d^2 + [a(M-1)]^2}}{2M} \qquad (4)$$

where d is the resolution of the detector, M is the amplification factor (the ratio of the distance from the X-ray source to the detector to the X-ray source to the sample), and a is the focal spot size of the X-ray tube. This means that large magnification, small focal spot size, and high detector resolution contribute to the high spatial resolution of CT scans. In order to avoid damaging the anode of the X-ray tube and achieve fast scanning, a trade-off must be made between the size of the focal spot of the X-ray tube and the spatial resolution of the CT image, so the maximum spatial resolution of a macro CT (an X-ray tube with a focal spot size of 1 mm) is limited to about 100μm [23]. Due to this limitation, finer pores and cracks are difficult to detect in CT detection, which leads to a good analysis of microstructure using this method. For the purposes of detecting such microstructure and obtaining better reconstructed image quality with higher efficiency, improving the CT image acquisition speed and spatial resolution is the main research approach at present. Spatial resolution is defined as the size or size of the smallest unit that the instrument can distinguish in detail, and factors affecting spatial resolution include device factors such as projected focal spot size and detector element size. Focal spots are areas on the anode of an X-ray tube or accelerator target that are hit by electrons and emit X-rays. The focal spot size affects the spatial resolution of the image. In general, the fine spot size improves the spatial resolution. The size of the sensor, as the terminal for receiving X-rays, also determines the microscopic degree of spatial resolution. To solve this problem, the researchers decided to use synchrotron radiation and microfocus X-ray tubes as X-ray sources to solve this problem through device iteration [24, 25].

The two devices are broadly the same in principle, however laboratory equipment typically features multicolour and conical beams, while synchrotron X-ray beams are essentially parallel, monochromatic, and have a better advantage in brightness. These features affect image quality and acquisition time. The current generation of laboratory X-ray CT systems can achieve spatial resolution of the same order as synchrotron X-ray CT [26]. Laboratory CT systems are not only economical and more accessible, however also meet the needs of most researchers [27]. However, image quality depends on more than resolution [28] and is also affected by SNR [29], phase contrast [30] and the presence of artifacts [31]. The high signal-to-noise ratio, the phase contrast level provided, and the absence of beam hardening artifacts make synchrotron X-ray CT particularly suitable for low-contrast materials, such as carbon fiber reinforced polymer composites (CFRP). Laboratory Microfocus CT System are ideal for conducting experiments on composites with higher contrast, such as glass fiber reinforced polymer composites (GFRP). In addition, because laboratory X-ray CT systems typically use divergent conical beams, they can accommodate large components that allow multi-scale studies; In addition, the conical beam geometry means that you can very easily continuously change the resolution (within practical limits) to fit the sample size and focus on a specific ROI after the measurement scans a larger volume. One of the main advantages of using synchrotron X-ray CT is that rapid experiments can be performed [32].

The high throughput and brightness characteristics of the synchrotron beam allow users to acquire up to 20 tomography images per second [32, 33]，however the acquisition time of laboratory X-ray CT varies from a few minutes (low resolution~10-0.5 mm) to a few hours (high resolution~150-50nm). Notably, the spatial resolution of X-ray microscopes equipped with Fresnel zone plate (FZP) optics, commonly used as objectives in analytical microscopy applications, such as nano-CT, is now approaching about 10 nm, close to the theoretical limit of simple FZP optics [34]. However, further spatial resolution is not always advantageous due to the limited field of view (FOV), and too small a region of interest (ROI) cannot truly represent the sample [34, 35].

According to the classification of spatial resolution, commonly used tomography systems can be roughly divided into conventional Macro-CT, Micro-CT and Nano-CT which is shown in Figure 3. For example, a system with a focal spot size greater than 0.1mm is considered a conventional CT system, while a Micro-CT system has a pixel size as low as one or several microns. Nano-CT can achieve pixel sizes as low as 0.4 microns [36]. Micro-CT and systems with higher resolutions can theoretically detect micro-damage or cracks that are sufficiently small and can affect the structural strength of materials, meeting the experimental requirements for micro-damage research and have been confirmed in some research.

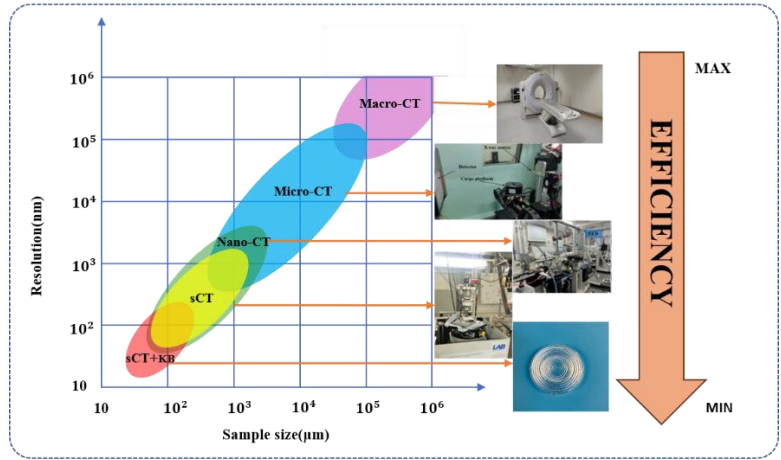

Figure 3: CT classification based on resolution and sample size.

### 2.3 Evolution and Applications of Micro-CT in Non-Destructive Testing and Material Characterization

In the 1980s, CT also gradually became popular in the field of NDT, which provides a new method for observing the structural damage inside and allows researchers to further understand the characteristics of materials. Thanks to the common development of X-ray source, sensor and computer, Micro-CT has been widely used in the micro field. Some researchers have also reported widespread use of Micro-CT based on a variety of research perspectives. In 2019, Sang-Yeop Chung published a review on the Micro-CT scanning of cement base materials. The author made a comprehensive analysis on the volume fraction and spatial distribution of pores and solid phases in cement-based materials with the help of Micro-CT, and introduced and analyzed the relationship between the above characteristics and the physical properties of cement-based materials [37]. Y. Wang published a review on the application and development of Micro-CT 3D analysis in the field of metallurgy in 2020. This review discussed the application of the Micro-CT system in geometry and mineral engineering. Through the three-dimensional analysis of Micro-CT, A number of mineral characteristics such as mineral permeability, mineral composition, mineral release and exposed particle surface area can be determined [36]. Yu Liu reviews quantitative evaluation methods for bone microstructure and bone formation based on Micro-CT, which has become a de facto tool for 3D non-destructive imaging of porous metal implants. At present, Micro-CT has been applied to the evaluation of microstructure and helps to explore the optimal microstructure of bone formation, thus providing a basis for the design of microstructure and the selection of materials [38].Qing Liu reviewed the research of Micro-CT in the field of shale microstructure, ranging from nano-pores to fractures of several microns to other rock physical properties. CT scans can capture the evolution of shale pore networks in certain experiments, such as pyrolysis, fracturing deformation, and chemical reactions, providing quantitative guidance for understanding the complex petrophysical changes occurring in shale reservoirs to optimize resource development [39].The research report sufficiently demonstrates that the method combining Micro-CT with volume rendering has been proven feasible for the detection of micro-damage. Furthermore, the integration with advanced visualization techniques can provide information that was previously unattainable, offering researchers reliable research tools and a broader range of viable research perspectives.

### 2.4 Strategies of Combating Noise and Artifacts in CT Imaging for Enhanced Image Quality

Due to the limitation of X-ray sources, there is always inherent system noise in CT images, which requires accurate image post-processing technology. Noise and various artifacts are common problems to be considered in the process of CT image capture.

#### 2.4.1 Balancing Spatial Resolution and Quantum Noise

In general, two types of noise must be considered in the projection data of CT-reconstructed images: additive noise and quantum noise. Additive noise refers to errors caused by electrical noise or rounding errors, which can be modeled as simple additive noise. Quantum noise is related to the number of X-ray photons emitted from the object, making error analysis more complex. The key to reducing image noise lies in the trade-off between spatial resolution and noise levels. Increasing resolution often involves detecting fewer primary photons, which can result in a blurrier image and consequently, reduced noise. Researchers can select operating parameters based on their specific needs [40]. Additionally, plane correction can be employed to reduce quantum noise in images by eliminating non-zero values[41]. Finally, iterative algorithm is the most widely used noise reduction techniques, significantly improving image quality compared to FBP [42]. These iterative algorithms include, but are not limited to ART, SART and SIRT [43]. Another effective denoising approach is the image space denoising algorithm, which applies linear filters or nonlinear filters like non-local mean (NLM) filtering directly to the reconstructed image [44].

### 2.4.2   Understanding and Mitigating Ring Artifacts

Ring artifacts arise from faulty detector elements, such as gain errors at specific locations within the detector array [45]. Ring artifacts show up as concentric rings on a CT image, particularly noticeable in a three-dimensional reconstruction. These artifacts are typically caused by variations in how individual pixels respond during the scanning process, which can be influenced by factors like temperature instability and radiation damage at varying scintillator thicknesses. While manufacturing defects in the detector can also lead to ring artifacts, this is rare in modern scanners [46]. Mismanagement and miscalibration of detector pixels, as well as impurities on the scintillation screen, can lead to ring artifacts that appear as circles centered around the axis of rotation. While all reconstructed slices contain these rings, their prominence varies depending on the slice's position relative to the axis of rotation. As a slice gets closer to the axis, the rings become more noticeable. The intensity of the ring artifacts also differs based on their source. Defective detector elements create sharp rings that are one pixel wide in the reconstruction. In contrast, a misaligned detector element results in a wider, less pronounced ring. Dust and damaged areas on the scintillator cause the broadest ring artifacts since they affect multiple detector pixels simultaneously. Throughout the scanning process, these ring artifacts can either overestimate or underestimate the attenuation values in the reconstructed images, complicating noise reduction and image segmentation. This makes quantitative analysis significantly challenging. To eliminate the influence of ring effects on the processing and quantitative analysis of reconstructed images, various algorithms can be employed, including both preprocessing and post-processing techniques. A commonly used preprocessing method is flat field correction. This traditional approach relies on the linear system response and involves correcting the original unfiltered image by subtracting it from the flat field image (also referred to as the white image), which is captured without any objects positioned between the X-ray source and the detector. Additionally, some preprocessing algorithms are applied to the projected data (sinusoidal graphs) to help minimize line distortion [45, 47]. The post-processing algorithm is applied directly to the reconstructed image. These methods utilize median, mean, and Gaussian filtering to compute the polar domain of reconstructed CT images, resulting in improved outcomes [48]. To optimize the performance of the algorithm that removes the ring effect, distortion can be addressed by implementing a combination of pre-processing and post-processing techniques applied to the image structure [49].

### 2.4.3   Addressing Beam Hardening Effects

Certain substances absorb lower energy X-rays more effectively than higher energy X-rays. As a result, the heterogeneous X-ray beam that passes through the absorbing medium contains a higher proportion of high-energy photons, making it more penetrating or "harder." This principle can lead to density or composition gradients that are not actually present in the imaged object, a phenomenon known as the "beam hardening effect" [50]. Currently, the primary method to eliminate beam hardening is to incorporate a beam hardening correction algorithm alongside the iterative algorithm, allowing for the correction of beam hardening artifacts during both image acquisition and reconstruction [50, 51]. Another method for correction is the empirical method, which relies less on prior knowledge [52].

### 2.4.4   Addressing the Partial Volume Effect

Partial volume effect (PVE) is a significant artifact in CT imaging that can greatly impact quantitative analysis. This effect within different image regions primarily arises from the system's point spread function (PSF). Additionally, there is another type of PVE related to the voxel size used in imaging, which can also affect the sampling process [53]. The value of each voxel reflects the average attenuation coefficient of all materials present within it. When a voxel contains two or more phases, the data acquisition process incorporates the respective attenuation coefficients. Consequently, the voxel values represent a linear combination of these coefficients, which is influenced by the volume each phase occupies within the voxel. As a result, the attenuation coefficient value may pertain to a material that is absent. The degree of error is contingent upon the resolution or size of the voxel. Consequently, boundaries in the resulting image may appear "stepped" or exhibit uneven pixel intensity levels. A variety of methods have been proposed for partial volume correction (PVC), which can be categorized into two main types: in-reconstruction methods and post-reconstruction methods, including projection-based techniques such as partial projection methods [54]. Various publications discussing the point spread function of scanner systems, known as deconvolution, propose a new correction method that incorporates iterative deconvolution and minor clipping. This method not only reduces noise in the image but also addresses some of the volume effect [55]. The researchers employed a 3D maximum likelihood expectation maximization algorithm for post-reconstruction partial volume effect correction through iterative deconvolution, thereby enhancing the accuracy of activity and contrast recovery [56].

### 2.4.5   Managing Scattering Artifacts

Scattering is a significant issue in conical beam computed tomography (CBCT) imaging. Modern flat panel detectors can capture complete two-dimensional data instantaneously while maintaining high isotropic spatial resolution. While they expand the volume coverage, they also increase the contribution of scattered radiation to the total detected signal. When back projecting this overestimated intensity, we inadvertently exaggerate the intensity of each voxel along the path. This leads to an underestimation of absorption. The reconstruction error varies depending on the object and is directly proportional to the amount of scattering present [57]. Scattering management strategies can generally be categorized into two main approaches: (1) devices designed to limit the number of scattered photons that reach the detector, such as backscattering grids; and (2) post-scattering correction or subtraction techniques. While the first method can effectively reduce the amount of detected scattered radiation, it does not eliminate completely. Notably, employing grids may impose certain physical constraints on the scanner's geometry and could require higher radiation exposures [58].

*2.4.6   Summary of Parameters and sample selections for high precision dataset*

Table 2: Summary of artifacts that occur with a Micro-CT system

| Type of artifact | Example | Source | Solution | Reference |
|---|---|---|---|---|
| Noise | 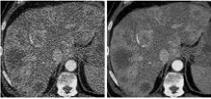 | Physical artifact; Inconsistent. | Binning data (reduce spatial resolution); Flat-field correction Iterative reconstruction; Filtering. | [21, 44] |
| Beam hardening | 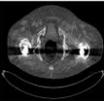 | Scanning artifact; Deviation of the detectors. | Recalibration of the detectors; Digital filtering. | [21, 52, 57] |
| Ring artifact | 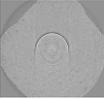 | Physical artifact; Unequal absorption of photons in the polychromatic X-ray beam. | Digital filtering; Calibration correction; Beam hardening correction software; Operator experience to select appropriate scan field of view. | [45, 57] |
| Partial volume effect | 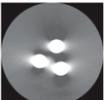 | Physical artifact; Voxel comprised of several phases; yielding average CT values of those phases. | Interpolation; Using higher spatial resolution. | [59] |
| Scattering artifacts | 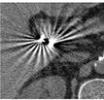 | Physical artifact; Diffracted photons. | Anti-scatter grid; Posteriori correction/ subtraction of scatter. | [57, 60, 61] |

In conclusion, due to the limitations of laboratory X-ray sources, CT data acquisition often includes inherent noise and artifacts. These issues result from the physical interactions between the material and the X-ray beam, as well as from the scanning process involving the detector. A summary of these artifacts is presented in Table 2. To minimize image artifacts and obtain high-quality volume data, it is essential for CT researchers to choose appropriate scanning parameters and sample selections to obtain high precision dataset.

## 3.  ADVANCEMENTS IN CT RECONSTRUCTION ALGORITHMS FROM ANALYTICAL TO DEEP LEARNING METHODS

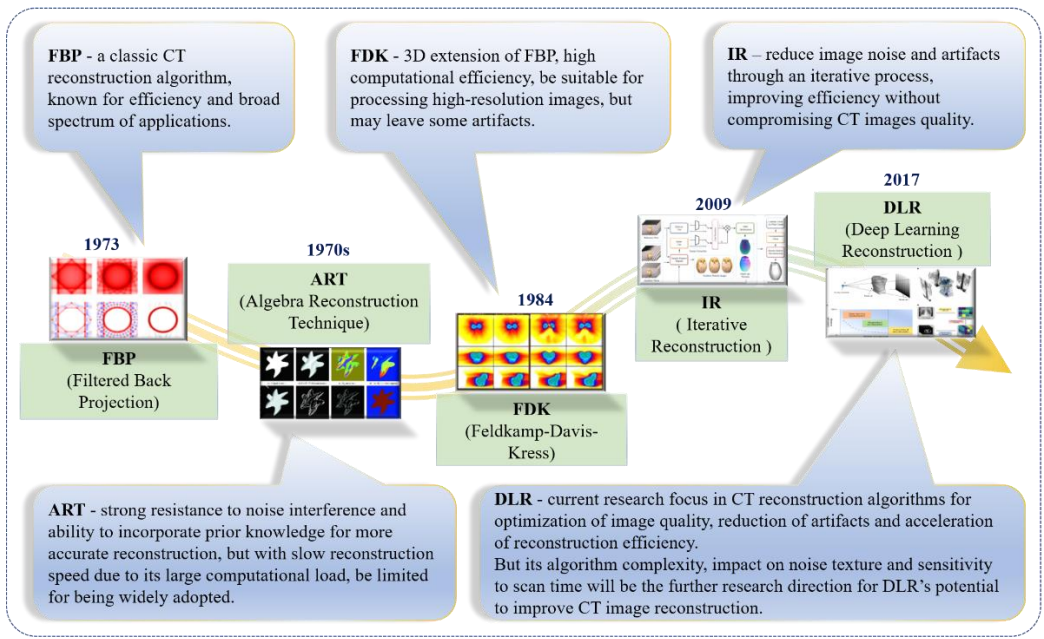

Figure 4: Development history of CT reconstruction algorithms.

There are two traditional types of reconstruction algorithms: analytical algorithms and iterative algorithms. Analytic reconstruction methods are efficient (fast) and elegant, however they cannot handle complex factors such as scattering. Filtered back projection (FBP) is a commonly used image reconstruction technique, although its efficiency is low. Iterative reconstruction (IR) can improve CT image quality by reducing quantum noise and artifacts and can realize more quantitative reconstruction based on optimization theory. In addition, with the rapid development of artificial intelligence, deep learning-based reconstruction has gradually become a hot topic in the field of reconstruction algorithm research and has been rapidly evolving. It has also gained a certain level of usability in practical applications. The development history of reconstruction algorithms is shown in Figure 4.

## 3.1 Analytic Reconstruction Algorithm and Their Fast Reconstruction in CT Image

FBP is a classical CT reconstruction algorithm. FBP reduces image blur in CT images by simple back projection as shown in Figure 5. A simple back projection smears each view along the path of the acquired image to form an image, resulting in a blurry version of the correct image. The final image is obtained by setting all pixel values along the line of light to the same value and summing the back projection view. To correct the blurring effect of simple back projection, FBP is used. Before back projection, each view is filtered by convolution with the filter kernel to create a filtered set of views. Through back projection of the filtered image, the obtained image can better represent the correct image, and several new filtering methods are proposed, such as Hamming filtering, Shep-Logan filtering, Ramachand Ran-Lakshminarayanan filtering. In addition, FBP can use fast Fourier transform (FFT), which enables fast reconstruction [62].

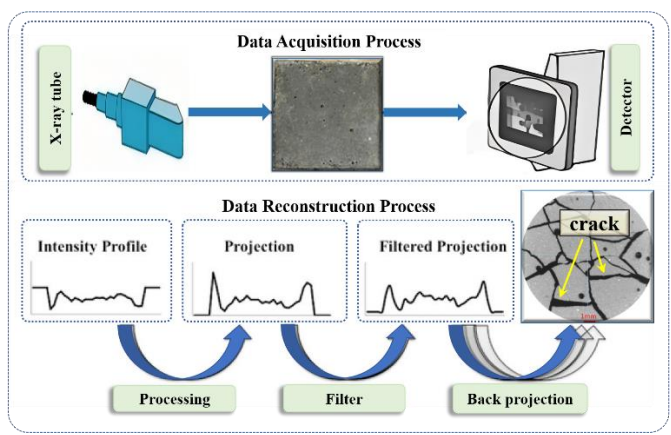

Figure 5: Principle and process of FBP.

In Micro-CT, cone-beam scanning has a larger projection range and higher photon efficiency than fan-beam scanning, so it is a common scanning method. Feldkamp et al. [63] proposed a three-dimensional analytic reconstruction algorithm for circular trajectory cone-beam scanning in 1984, namely FDK algorithm. In the orbital plane, it is equivalent to FBP reconstruction of Fan-beam, while in the area far away from the orbital plane, the reconstructed image will have obvious cone-beam artifacts, which are mainly manifested as low image value and cross-string images between layers. Therefore, FDK algorithm is only suitable for small cone Angle scanning reconstruction, and can only accurately reconstruct the orbital plane or the longitudinal uniform object. There are three main ways to improve FDK algorithm: (1) Add correction items to slow down the longitudinal attenuation of reconstruction results. Hu [64] proposed an improved FDK algorithm, which regarded the original image as the sum of three terms: FDK reconstruction results, FDK reconstruction unused cone-beam CT scan information of circular trajectory and missing data of circular trajectory scan. He used the inverse Radon transform to calculate the second term and pointed out that the third term could be estimated by prior knowledge to make the reconstruction results more accurate. On this basis, Zhu et al. [65] used Radon spatial interpolation and parallel beam back projection to correct the reconstruction results. (2) Data rearrangement：P-FDK (parallel FDK) algorithm [66] performs a similar fan-beam rearrangement process on the projection data in the horizontal direction, while keeping the longitudinal direction unchanged to obtain a "Cone-Parallel" projection, which makes the central virtual detector bend backward and improves the reconstruction speed. On the basis of P-FDK,T-FDK (tent FDK) algorithm [67] also interpolates in the longitudinal direction, changes the filtering path, and can significantly improve the reconstruction error under the large cone Angle. Based on T-FDK,C-FDK (curve-filtered FDK) algorithm [68] selects different filtered curves on the central virtual detector plane to further improve the reconstruction quality. (3) Projection weighting, Based on the "Cone-Parallel" projection, Tang et al. [69] proposed a 3D weighting strategy based on the properties of conjugate rays, which can effectively alleviate the cone-angle artifacts in reconstructed images.

## 3.2 Iterative Reconstruction Algorithms Overcoming FBP Limitations and Enhancing CT Image Quality

At the same time, In order to overcome the considerable image noise generated when the speed is reduced in the process of FBP image reconstruction [70], IR is designed to reduce image noise, and the detail boundary is smoother than FBP [71, 72]. In recent years, mainstream CT system vendors have also introduced their own commercial IR, mainly divided into hybrid-IR and MBIR [43, 73, 74]. FBP involves filtering the projection data, or sinograms, with a high-pass filter and then back-projecting them into the image domain, as shown in the left column. In hybrid iterative reconstruction (HIR), sinograms are subjected to iterative filtering to minimize artifacts and are back-projected into the image domain. Following this, images undergo iterative filtering to reduce noise. In contrast, model-based IR begins with the back-projection of sinograms into the image domain. Utilizing models that simulate the acquisition process, images are then forward-projected into the sinogram domain to create artificial sinograms. These artificial sinograms are compared with the actual sinograms, allowing for the optimization of images and the removal of noise[75]. The flowcharts of these algorithms are shown in Figure 6.

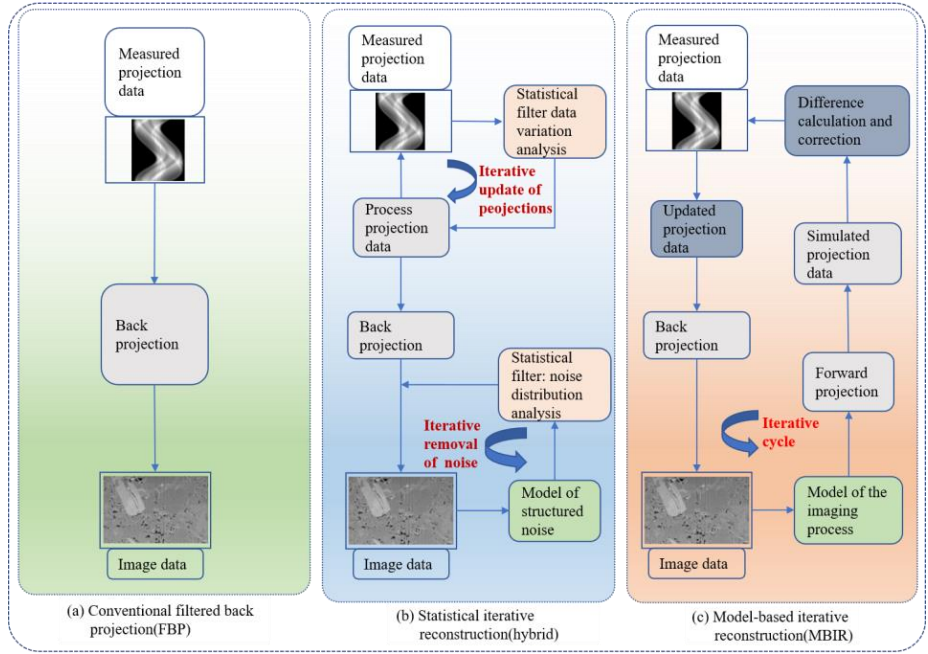

Figure 6: Comparison of the principle of FBP, Hybrid-IR and MBIR

Veo is the first fully iterative IR algorithm (model-based iterative algorithm) in clinical practice and one of the most advanced algorithms to date [76]. In full IR, the raw data is back-projected into the cross-sectional image space. Subsequently, the spatial data of the image is projected forward to calculate the artificial raw data. The forward projection step is the core block of the IR algorithm because it enables physically correct modulation of the data acquisition process, including the system geometry and noise model. The section image is updated by comparing the artificial raw data with the real raw data. At the same time, the image noise is removed through the regularization step. The process of backward and forward projection is repeated until the difference between the real raw data and the artificial raw data is minimized. As can be imagined, full IR is computationally more demanding than Hybrid-IR, resulting in a longer reconstruction time for full IR. Because of the long rebuild time, the supplier decided to develop a different advanced algorithm called ASIR-V (GE Healthcare), which was approved by the FDA in 2014. Meanwhile, other vendors have introduced other hybrid and model-based IR algorithms, including AIDR3D (Adaptive Iterative Dose Reduction 3D, Canon Healthcare), ADAMIRE (Advanced Modelling Iterative Reconstruction, Siemens Healthcare), and IMR (Iterative Model reconstruction, Philips Healthcare). Most recently, in 2016, the model-based IR algorithm FIRST (Iterative reconstruction Solution based on forward projection Model, Canon Healthcare) received FDA approval. Table 3 shows the advantages and disadvantages of the commercial mainstream algorithms. Overall, these studies show that the image quality and diagnostic value of IR are improved compared to FBP. The IR radiation dose can be reduced by 23% to 76% without compromising image quality [77]. Several studies have compared different approaches from multiple vendors, and in general, these studies have found that model-based IR can further reduce radiation dose compared to hybrid IR and FBP [78]. In a systematic review of 24 studies, Den Harder et al. [79] found that the average radiation dose of a reconstructed chest CT scan using FBP could be reduced to 1.4(0.7-7.8)mSv by applying IR. However, reducing the CT radiation dose to low-contrast areas of the body, such as the abdomen, is more problematic [80]. At lower radiation doses, IR does not always improve the detectability of low-contrast lesions [81]. However, most studies have found that IR does reduce the radiation dose at high contrast locations without compromising image quality [82, 83]. The relatively small amount of radiation also means shorter exposure times and better CT scan efficiency.

Table 3: The advantages and disadvantages of the commercial mainstream algorithms of the four major suppliers

| Vendor | Algorithm name | Type of algorithm | Reconstruction speed | Artifact reduction | Noise reduction |
|---|---|---|---|---|---|
| GE Healthcare | ASIR(Adaptive Statistical Iterative Reconstruction) | Hybrid | + | + | ++ |
| | Veo(MBIR) | Model-based | - | ++ | +++ |
| | ASIR-V | Hybrid | + | + | ++ |
| Philips Healthcare | iDose[4] | Hybrid | + | + | ++ |
| | IMR(iterative model reconstruction) | Model-based | - | ++ | +++ |
| Siemens Healthcare | ISIR(iterative reconstruction in image space) | Image domain | ++ | - | + |
| | SAFIRE(sinogram-affirmed iterative reconstruction) | Hybrid | + | + | ++ |
| | ADMIRE(advanced iterative dose reduction 3D) | Model-based | - | ++ | +++ |
| Canon Healthcare | AIDR3D(adaptive iterative dose reduction 3D) | Hybrid | + | + | ++ |
| | FIRST(forward projected model-based iterative reconstruction solution) | Model-based | - | ++ | +++ |

### 3.3 Deep Learning in CT Image Reconstruction from Traditional Algorithms to AI-Enhanced Techniques

With the rise of artificial intelligence. In addition to image classification, object detection and playing games [84, 85], Ai has also gained significant interest for its potential to improve CT image reconstruction [86]. Artificial intelligence, and more specifically machine learning, is a set of mappings capable of moving from raw inputs (such as the intensity of individual pixels) to specific outputs (such as the classification of diseases) [87]. Multiple research groups are working on applying artificial intelligence to improve the reconstruction of CT images. One application is image-space based reconstruction, where a convolutional neural network is trained on low-dose CT images to reconstruct conventional dose CT images [88-90].

Over the past few years, researchers have developed an alternative to these traditional refactoring approaches, called the Deep Learning Reconstruction (DLR) approach [91]. In general, IR algorithms are based on a prior function designed by humans to obtain a low noise image without loss of structure [75, 92-94]. DLR allow for more complex functions, which have the potential to enable low-dose CT [95-97] and sparsely sampled CT [98]. These AI technologies have the potential to reduce CT radiation doses and optimize image quality while speeding up reconstruction times. Several DLT have been reported in the last five years [99, 92, 100]. These techniques fall into two groups: direct DLR frameworks and indirect DLR frameworks. These categories are distinguished based on the use of FBP or IR.

In indirect DLR frameworks, either FBP or IR is used. The three types of indirect frameworks are differentiated based on when the deep learning network is deployed. Sinogram-based frameworks prioritize the optimization of the sinogram and position the network before the sinogram undergoes FBP or IR. Conversely, image-based frameworks utilize the network to refine the image following the initial reconstruction via FBP or IR. Hybrid frameworks integrate both sinogram and image optimization techniques. In indirect reconstruction, deep learning networks predominantly rely on convolutional neural networks (CNNs). Two examples are wavelet transform–based U-Net [95] and residual encoder-decoder CNN [89]. Transformer-based neural networks and generative adversarial networks [86] are examples of a non-CNN–based approach to image optimization [101].

Without the need of FBP or IR, direct DLR algorithms recreate the sinogram straight into a picture. This may lessen artifacts brought about by IR or FBP. This is only feasible, though, provided there are no FBP- or IR-related artifacts in the ground truth photos that were used to train the model. The direct algorithms AUTOMAP and iRadonMAP are two examples. [102]

So far, the U.S. Food and Drug Administration (FDA) has approved three DLR algorithms [103]. The first algorithm, called True Fidelity, was developed by GE Healthcare (GE Healthcare, Waukeha, WI, USA)[70], and trained on FBP images of high radiation doses obtained on patients and models. Canon's Advanced ClearIQ Engine (AICE), Boedeker K. AiCE DLR bringing the power of ultra-high resolution CT to routine imaging named Canon Medical Systems Corporation in 2019, is an indirect image-based DLR algorithm that is trained using high-quality IR patient images. A third commercially available DLR algorithm is the Precision Image (PI) recently produced by Philips (Philips Healthcare, Cleveland, Ohio, USA). PI trains convolutional neural networks (CNNS) to directly reconstruct low dose simulated sinusoidal graphs into images that reproduce the appearance of images of patients with conventional doses of FBP. The DLR algorithm generates high-quality images, although its input is low-dose acquisition, thus allowing for minimizing CT acquisition time, which has important implications for patient safety and acquisition efficiency [104, 105].

The NPS of DLR images is similar to FBP [104]. Using TrueFidelity from GE Healthcare to measure the average frequency of DLR images, meaningful shifts in DLR images relative to FBP were measured only at low scan times [94]. Canon Medical Systems' DLR solution (AiCE) showed NPS similar to MBIR [106]. Some studies have shown that the intensity (or weight) of DLR processing will affect the noise texture, and the currently available algorithms on the two scanners [107] will move to the low frequency at high DLR intensity, however some studies have shown that the new version of DLR has alleviated this change [108].

In the assessment of absolute image noise, the DLR has the same or better noise reduction level than MBIR and FBP [104]. The noise reduction capabilities of the DLR allow it to explore new spatial resolution limits, such as deploying detectors with smaller pixels without being affected by noise, and the DLR likewise shows high contrast spatial resolution comparable to MBR methods that produce similar results at 50% and 10% modulation transfer function (MTF) points [106]. While task-based MTF differs slightly at different tissue contrast levels of the DLR, the differences measured are comparable to the differences measured to the measurements, and overall the MTF values of the DLR are similar or better than those of the FBP [105]. This is unlike the IR, which has shown low MTF values relative to FBP across multiple vendor implementations, especially in low-contrast targets [109, 110]. It has been observed that DLR's task-based MTF is stable over a certain range of scan times, and the spatial resolution remains the same as the scan time decreases [105],however there is a downward trend in tissues with very short scan times and low contrast [94]. Studies have shown that this phenomenon has been reduced in the new version of the DLR [108]. All in all, DLR methods described using model data do not show MTF declines at low contrast levels like IR methods.

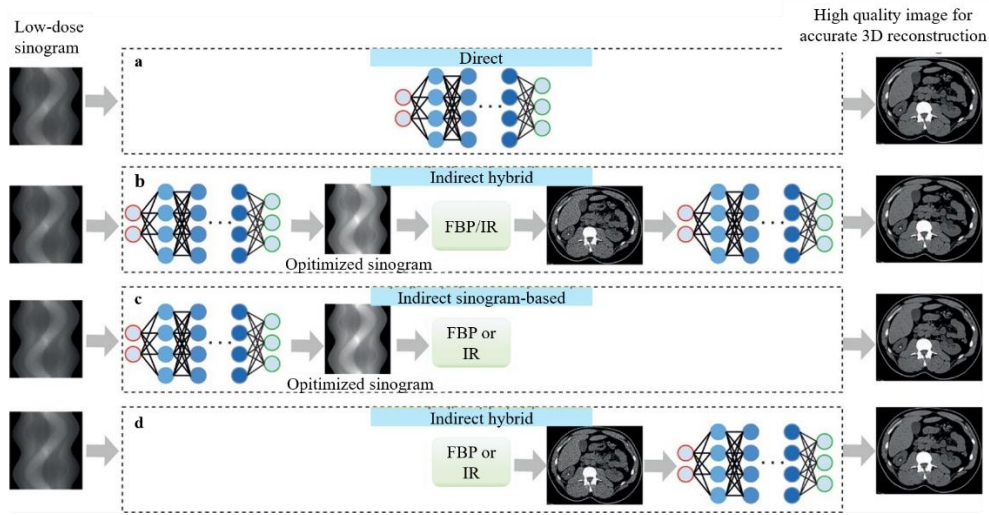

Figure 7: The diagram provides an overview of various types of deep learning reconstruction (DLR) methods. (a) Direct DLR algorithms reconstructure a high-quality image directly from the sinogram, bypassing both filtered back projection (FBP) and iterative reconstruction (IR). (b) The indirect hybrid method combines optimization in both the sinogram and image domains with an FBP or IR step to generate a high-quality image. (c) The indirect sinogram-based method focuses solely on optimizing the sinogram prior to reconstruction using FBP or IR. (d) The indirect image-based method first employs FBP or IR for initial image reconstruction, followed by optimization in the image domain. [75]

DLR offer faster reconstruction speeds, better dose-performance, and fewer noise and artifacts compared to FBP and IR. Figure 7 provides an overview of various types of DLR. Thanks to the rapid iterations of DLR, initial issues with DLR are being quickly resolved. However, DLR is not without its flaws. To achieve better reconstruction results, a large and appropriate training dataset as well as hardware computational power are required. Additionally, the reconstruction process may sometimes generate phantom artifacts that do not exist, which can affect the judgment of defects. Comparison of CT reconstruction algorithms is in Table 4.

Table 4: Comparison of CT reconstruction algorithms

| Reconstruction algorithms | Advantage | Disadvantage |
|---|---|---|
| Filtered Back Projection (FBP) | Reconstruction speed<br>Image texture | Image noise<br>Artifact<br>Radiation dose |
| Hybrid Iterative Reconstruction (Hybrid-IR) | Noise reduction<br>Resolution of Low Contrast Objects<br>Reduced dose and artifacts (moderate) | Reconstruction speed<br>Transition smoothing<br>Image texture |
| Model-Based Iterative Reconstruction (MBIR) | Noise Reduction (Strong)<br>Resolution of Low Contrast Objects<br>Reduced dose and artifacts (strong) | Reconstruction speed<br>Transition smoothing<br>Image texture |
| Artificial Intelligence - Deep Learning Reconstruction | Noise Reduction (Strong)<br>Low contrast object resolution (strong)<br>Reduced dose and artifacts (strong)<br>Spatial resolution preserved (strong) | Training dataset selection<br>Size of the training set<br>Data enhancement |

## 4. ENHANCING VOLUME RENDERING TECHNIQUES IN ALGORITHMS, ACCELERATION, AND DATA REDUCTION

As shown in Figure 8, Ray Casting [111] is a volume rendering algorithm in order of image space. The basic principle is: From each pixel in the screen, a ray is projected into the volume data field, and this ray passes through the three-dimensional data field. Along this ray, k isometric sampling points are selected, and 8 is selected from the nearest sampling point. The color values of the data points are interpolated trilinear (resampling). In the last step of image synthesis, the color values and opacity values of each sampling point on each ray are synthesized from front to back or from back to front, that is, the color values of the pixel points emitting the ray are obtained. Resampling and image synthesis are performed on a per-pixel basis for each scan line on the screen.

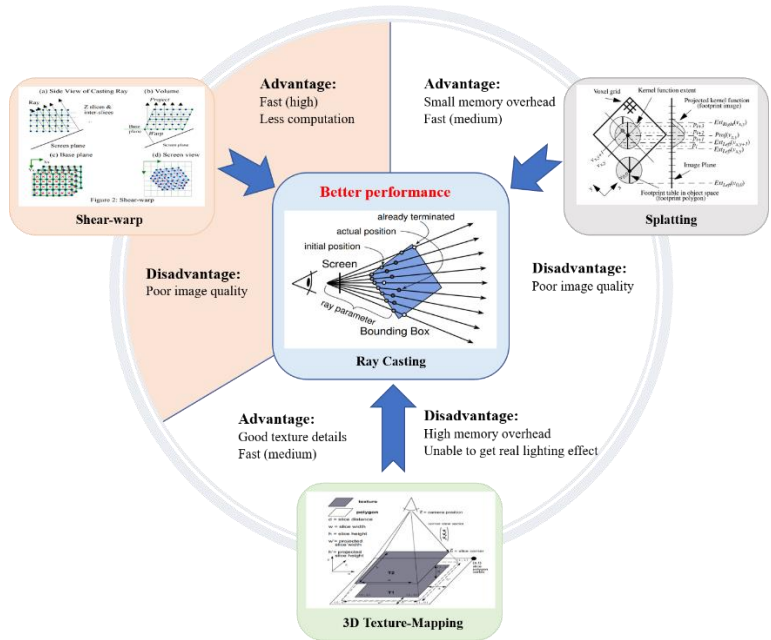

Figure 8: The advantages and disadvantages of four mainstream algorithms.

Ray Casting is a traditional volume rendering method, which has significant advantages and disadvantages. The advantage is that the quality of the drawing is quite high, the imaging is clear, and the details of the object can be better reflected. The only drawback is that due to the use of all voxels, the calculation amount of data is particularly large, and the memory usage is relatively high, which reduces the speed of volume rendering to a certain extent and cannot meet the real-time interaction needs of users. Common volume rendering include Splatting [112], shear-warp [113], 3D Texture-Mapping [114]. These algorithms were essentially compromises made in the era of limited computational power, where volume rendering quality was sacrificed for interactivity and rendering speed. However, with the enhancement of computational power in the present day, researchers now favor ray-casting methods that offer higher rendering quality and better detail representation.

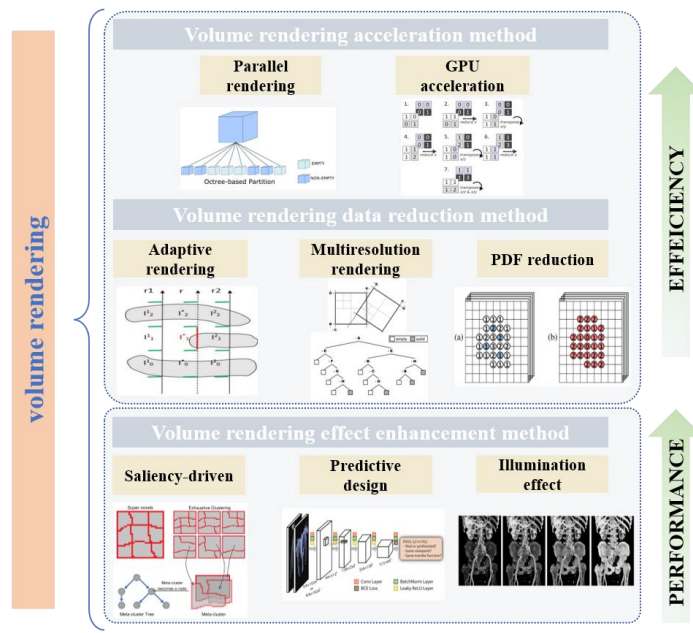

Figure 9: Various method to improve the efficiency and performance of ray casting.

Ray casting method has excellent rendering quality that other method can't get, so many researchers have done a lot of work on the acceleration of this algorithm as shown in Figure 9, researchers have conducted in-depth research on the volume rendering algorithm from many aspects to continuously improve the rendering efficiency and enhance the rendering effect. Based on the different technical approaches and principles used by researchers for algorithm optimization, they can be mainly categorized into three methods: volume rendering acceleration, data simplification for volume rendering, and enhancement of volume rendering effects. The first method primarily achieves faster rendering speeds by better scheduling and utilization of hardware resources such as CPUs and GPUs. The second method improves rendering efficiency by simplifying the data that needs to be rendered. The third method mainly enhances visual effects by strengthening the boundaries and lighting effects of rendered objects. The main purpose of volume rendering acceleration and data simplification for volume rendering is to improve efficiency and another one is to enhance the effect. These three optimization approaches will be specifically introduced in Chapter 4.

## 4.1 Volume Rendering Acceleration

### 4.1.1 Parallel Volume Rendering Efficiency

Parallel volume rendering algorithm is an effective way to improve rendering efficiency. Various volume rendering algorithms can improve performance through the multi-core acceleration capability of parallel machines. Because the main computation of volume rendering is concentrated in two stages: sampling calculation and image synthesis, the typical parallel mode can be divided into two steps: data parallel and image parallel. and data organization and load balancing are important factors affecting the parallel efficiency.

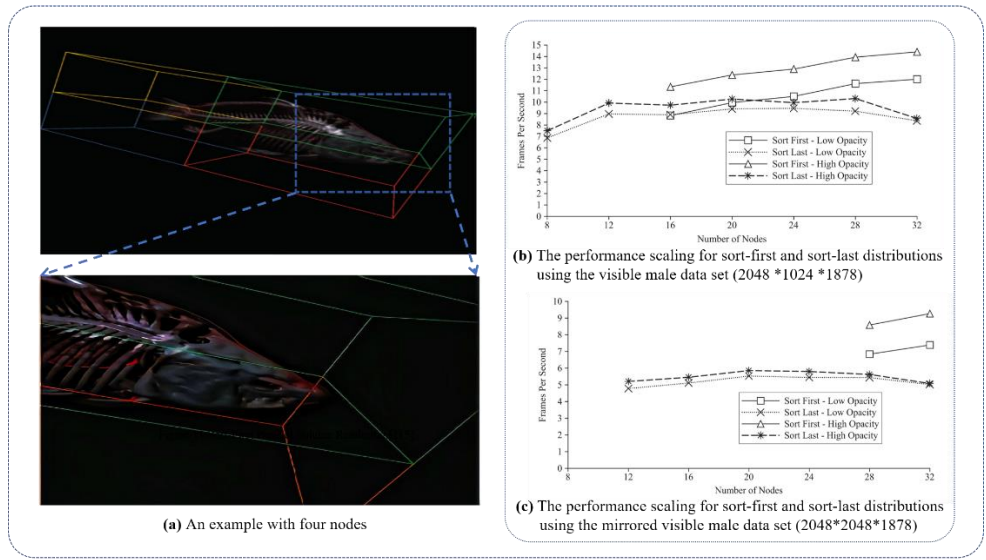

**(a)** An example with four nodes

**(b)** The performance scaling for sort-first and sort-last distributions using the visible male data set (2048 *1024 *1878)

**(c)** The performance scaling for sort-first and sort-last distributions using the mirrored visible male data set (2048*2048*1878)

Figure 10: Sort-first parallel volume rendering. [115]

Moloney et al [115]present a sort-first parallel volume rendering algorithm which have shown in Figure 10 that data-scalable sort-first distributions can outperform traditional sort-last distributions in many scenarios. their proximity caching algorithm reduces spikes in loading and could reduce the loading overhead when asynchronous transfers are possible. Figure 12: Parallel distributed, GPU-accelerated, advanced lighting calculations [121] shows an example of a sort-last (object-space) and the performance scaling for sort-first and sort-last distributions using the different data set. Howison et al. [116] proposed MPI + OpenMP hybrid parallel rendering algorithm , which made full use of shared memory on multi-core nodes to reduce memory consumption and data communication, and the parallel rendering scale reached 21.6 million cores, compared to 46,083 cores The ultra-large scale grid data achieves interactive drawing frames of high-resolution images. Figure 11(a) shows the principles of the algorithm and Figure 11(b) shows the effect of the algorithm. The ray-casting speedup is linear with distributed memory, however sublinear with hybrid memory caused by the difference in decomposition geometries (cubic versus rectangular). Monte et al. [117] proposed a volume rendering algorithm, PD-Render intra for individual networked nodes in a parallel distributed architecture with a single GPU per node. The implemented algorithm can achieve photorealistic rendering as well as a high signal-to-noise ratio at interactive frame rates. Given the computing industry trend of increasing processing capacity by adding more cores to a chip, the focus of Bethel and Howison's [118] work is tuning the performance of a staple visualization algorithm, ray casting volume rendering, for shared memory parallelism on multi-core CPUs and many-core GPUs. Their results indicate the optimal configurations on the GPU occur at a crossover point between those that maintain good cache utilization and those that saturate computational throughput. Wang et al. [119] proposed a parallel rendering algorithm based on hybrid parallelization of both image and data space. In contrast to traditional parallel rendering algorithms, this algorithm totally removes image compositing stage by generating ray-data intersecting list for each pixel, and effectively improves the efficiency of volume rendering in in situ visualization scenarios without interaction. However, this algorithm is not fast enough when applied to very large datasets. Leaf et al. [120] presented a GPU-based parallel volume renderer for large scale AMR datasets. A partitioner for AMR datasets, and described a novel method for decomposing blocks by interpolation region to improve rendering effect. Shih et al. [121] presented a novel advanced lighting technique for sort-last distributed DVR which can effectively avoid data exchange between compute nodes during the lighting computation. The method improves visual quality of the rendered images by providing more spatial cues, enabling a better understanding of the depth relationship and surface shape of the features within the data. Figure 12 show the capability of this method to enhance large-scale volume rendering.

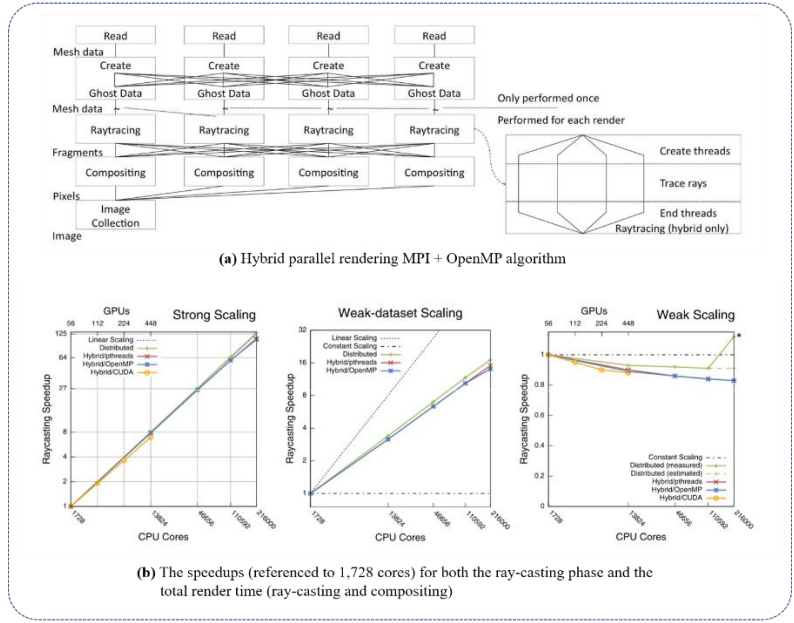

**(a)** Hybrid parallel rendering MPI + OpenMP algorithm

**(b)** The speedups (referenced to 1,728 cores) for both the ray-casting phase and the total render time (ray-casting and compositing)

Figure 11: Hybrid parallelism for volume rendering on large-, multi-, and many-core systems. [116]

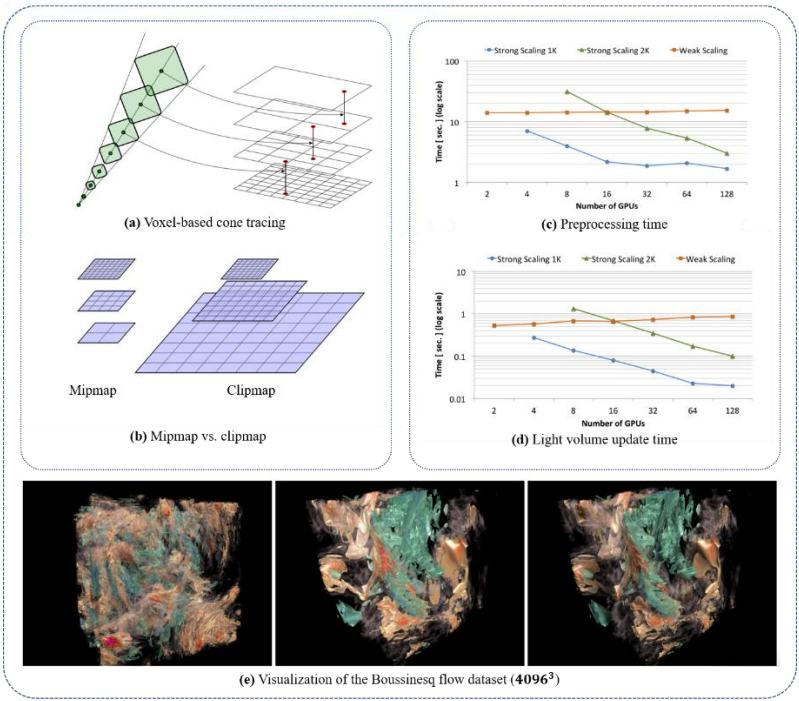

**(a)** Voxel-based cone tracing

**(b)** Mipmap vs. clipmap

**(c)** Preprocessing time

**(d)** Light volume update time

**(e)** Visualization of the Boussinesq flow dataset ($4096^3$)

Figure 12: Parallel distributed, GPU-accelerated, advanced lighting calculations [121]

### 4.1.2 GPU-Accelerated Volume Rendering

Using GPU to accelerate the sampling and synthesis calculation in the rendering process is another way to improve the efficiency of volume rendering. kD-Jump [122] is a recently proposed stackless traversal technique that can efficiently manipulate implicit kD trees so that the traversal directly returns to the next valid node without additional node access. On the other hand, Gpus are also often installed on visual parallel servers, so that people can use the mass storage capacity of parallel machines and the dual acceleration of multiple cpus coupled with multiple Gpus to deal with high quality and fast rendering of large-scale data sets. Zellmann et al. [123] presented two implementations of a parallel algorithm to fully rebuild k-d trees for spatial indexing of sparse volumes, one targeting multi-core CPU systems, and the other one targeting GPUs. The adapted GPU implementation employs binning, local AABB pre-calculation and z-order Morton curves for fast retrieval of neighboring blocks to optimize the recursive node splitting phase of the multi-core CPU algorithm. The GPU algorithm does not produce the exact same k-d trees as the CPU algorithm does, however enables far better scalability. Weiss and Westermann [124] introduced a framework for differentiable direct volume rendering (DiffDVR) which can automatically determine optimal parameter combinations regarding different loss functions to synthesize volume-rendered imagery for transfer learning tasks. Figure 13(d) shows the rendering results of DiffDVR and the comparison among DiffDVR, algebraic reconstruction provided by the ASTRA-toolbox and Mitsuba's differentiable path tracer. DiffDVR has the best results. Al-Ayyoub et al. [125] focused on the Fuzzy C-Means (FCM) clustering algorithm and employed GPU parallelism capabilities to improve its performance. The best implementation they devised to extract volume objects from the DICOM files was about 6.3× faster compared with the basic sequential implementation. Ströter et al. [126] presented a novel bounding volume hierarchy for GPU-accelerated DVR. The octree-based data structure is laid out linearly in memory using space filling Morton curves. This allows for significantly shorter construction times compared to the state of the art. For GPU-accelerated DVR, the new method achieves performance gain of 8.4× to 13×.

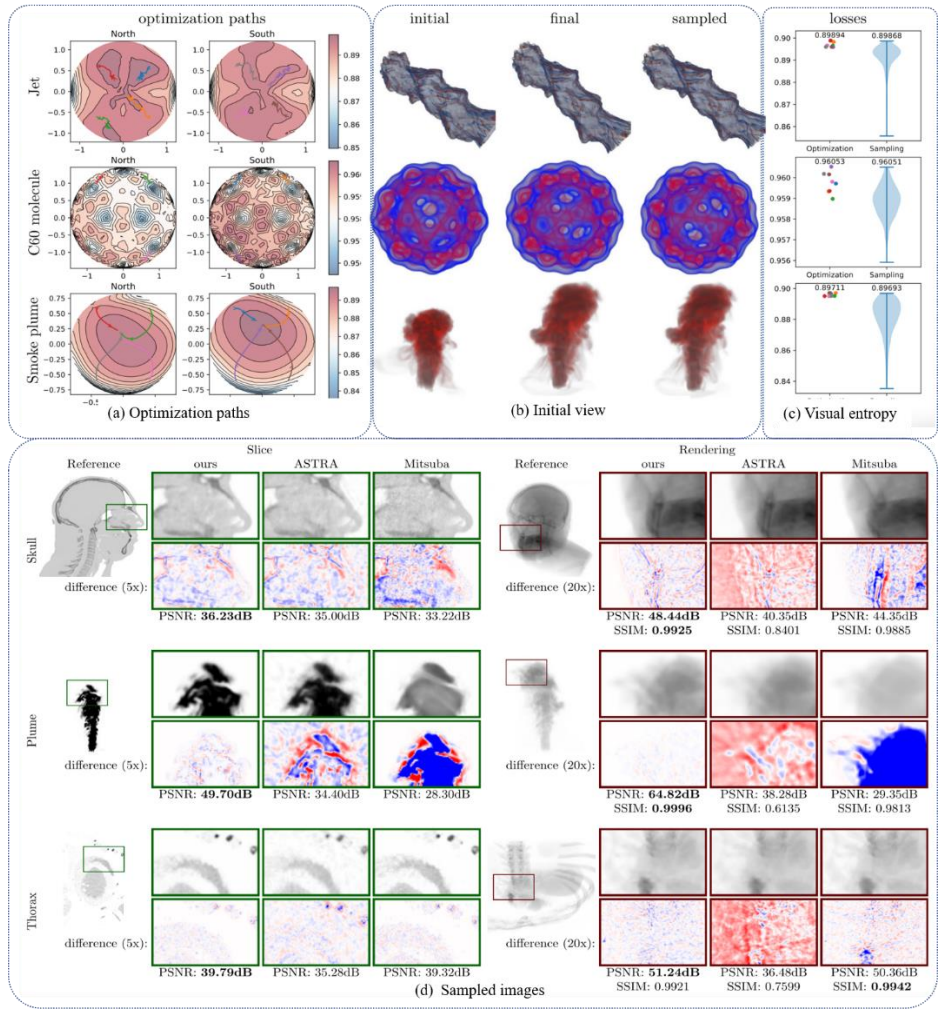

Figure 13: Differentiable direct volume rendering [124]. Best viewpoint selection using maximization of visual entropy for datasets jetstream (**256³**), potential field of a C60 molecule (**128³**), and smoke plume (**178³**). Comparison of DiffDVR with eight initializations against random uniform sampling over the sphere of 256 views. (a) Optimization paths over the sphere. (b) Initial view, selected view of DiffDVR, best sampled view. (c) Visual entropy of optimization results are colored points corresponding to (a). (d) Comparison among DiffDVR, ASTRA and Mitsuba.

### 4.1.3 Summary

Volume rendering acceleration is primarily achieved through parallel acceleration and GPU acceleration. Essentially, both methods aim to improve the efficiency and quality of volume rendering by better utilizing hardware performance. Parallel algorithms can take full advantage of the multi-core architecture of modern processors, distributing computational loads through data parallelism, and image parallelism processes multiple image blocks simultaneously to further enhance efficiency. The latest research has already reached a parallel computing scale of 21.6 million cores. Researchers achieve higher quality interactive real-time rendering by optimizing data organization and load balancing, reducing memory consumption, and minimizing data communication.

Nowadays, GPUs, with their tens of thousands of parallel processing cores (such as CUDA cores from NVIDIA), provide extremely high computational throughput. These cores are also optimized for graphics computing, allowing for more efficient execution of volume rendering tasks. Mainstream GPU-related algorithms can achieve a 6-14 times efficiency improvement. However, both acceleration methods have certain limitations. As the upper limits of semiconductor processes approach, it signals that the rapid development of hardware over the past few decades is about to hit a bottleneck. Researchers cannot solely rely on hardware performance improvements to enhance volume rendering efficiency. They need to combine CPU and GPU to achieve heterogeneous computing, making better use of the CPU's large memory for handling large datasets and the GPU's more powerful parallel computing capabilities. Algorithms should be optimized and fine-tuned to achieve better volume rendering effects.Additionally, technologies such as cloud computing and machine learning can be integrated to address the high hardware costs associated with these technologies, further enhancing the performance and application range of volume rendering.

## 4.2 Enhancing Volume Rendering Efficiency and Quality

### 4.2.1 Adaptive Sampling for Efficient Volume Rendering

The adaptive sampling method is used to depict the internal physical changes of data with fewer sampling points, thus reducing the overhead of computation, memory and communication, which is also an important way to improve the rendering efficiency. Common adaptive sampling methods include space leaping [127] , hierarchical adaptive sampling [128] , detail-oriented sampling [129] and gradient field magnitude method [130]. ISuwelack et al. [131] derived the appropriate sampling criteria from the conversion function and spectral decomposition of the dataset, and integrated a GPU-based adaptive raycast rendering algorithm. Corcoran and Dingliana [132] use the consistency between image frames to quickly generate two-dimensional importance maps, thus guiding adaptive sampling and realizing rapid refresh of volume rendering images when the viewpoint or conversion function changes. However, it is difficult for the above methods to accurately grasp the change rule of the value of physical quantities while reducing the sampling points. To solve this problem, Marchesin and Verdier [133] proposed an adaptive unit projection volume rendering technique, which can well grasp the monochromatic interval of the change of physical quantities in the rendering and obtain satisfactory rendering effects. The amount of data in large-scale scientific computing far exceeds the capacity of single hard disk and memory. At this stage, parallel processing volume rendering is indispensable. However, adaptive sampling in parallel mode will face difficulties in data organization of sampling points, spatial relationship of data blocks, ordering by viewpoint and load balancing. In order to adapt to large-scale scientific data, Wang et al. [134] improved Marchesin and Verdiere's adaptive sampling method. The algorithm is parallelized in a distributed and parallel environment, and the problems of sampling point management and load balancing are solved. Compared with the traditional uniform sampling method in a distributed environment, the performance of the algorithm is improved greatly. Based on the low efficiency of the sampling point selection and slow rendering speed of the existing ray-casting algorithm, Wang et al.[135] proposed a new method based on dynamic sampling and improved rendering operator. This method can effectively shorten the time of sampling and rendering and improve the efficiency of the algorithm on the premise of ensuring the high quality of graphics rendering as shown in Figure 14.

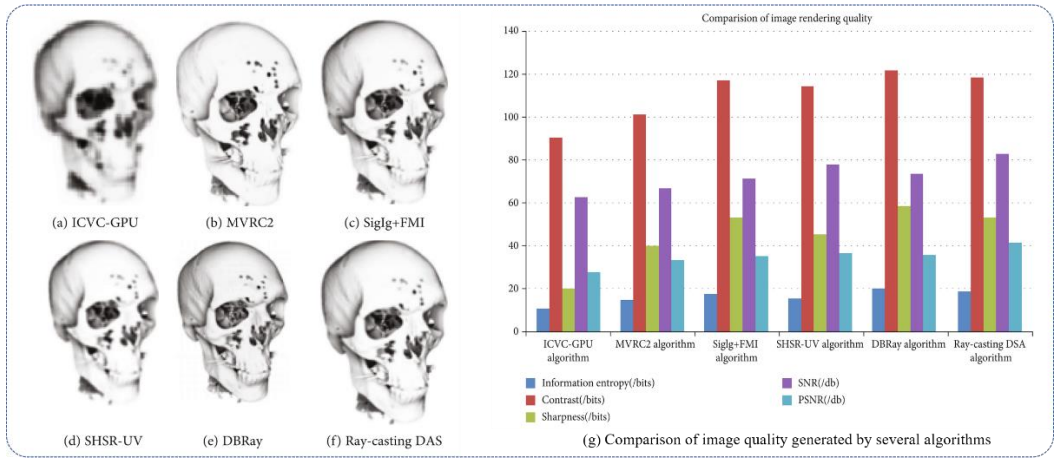

Figure 14: Ray-Casting Algorithm Using Dynamic Adaptive Sampling [135]. (a)-(f) Comparison of the visual effects of the different algorithms in image rendering. (g) Comparison of image quality generated by several algorithms.

Xue et al. [136] proposed a new semi-adaptive partitioning method and two efficient out-of-core volume rendering algorithms for the visualization of very large 3D image datasets. Compared with the methods based on large-scale parallel computers or in distributed environments, these methods can efficiently run on consumer PC hardware, which makes the application cost much lower and make result better when high-resolution and precise results are required as shown in Figure 15.

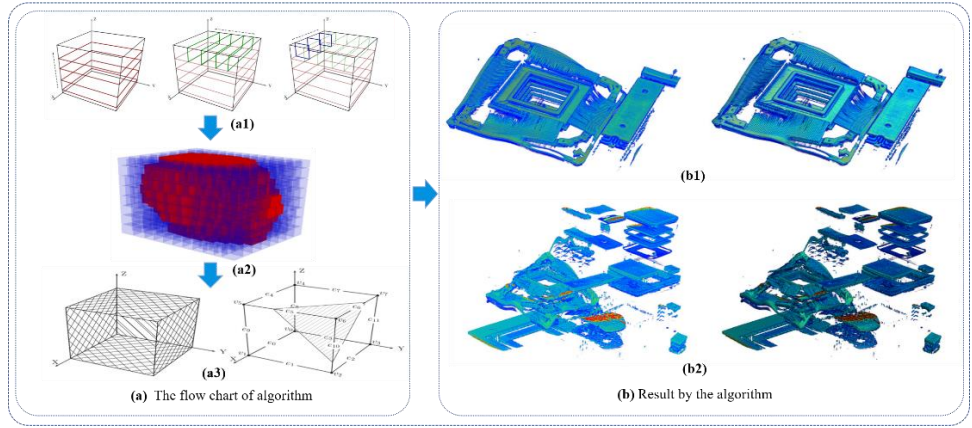

Figure 15: Volume rendering methods by semi-adaptive partitioning. (a) The flow chart of algorithm. (a1) Semi-adaptive partitioning. (a2) Semi-adaptive partitioning results. (a3) Generating texture polygons from the planes cutting through the bounding box of the 3D texture. (b) Result by the algorithm [136].

### 4.2.2 Balancing Image Quality and Rendering Speed in Multi-Resolution Volume Rendering

The development of multi-resolution rendering can compromise between image quality and rendering speed, so as to obtain the fastest rendering results under the limited resource budget, which is also a kind of method to deal with large-scale data and optimize rendering effect. [137] proposed a multi-resolution beam tracking volume rendering algorithm with drawing large volume data on ordinary PC. Suter et al. set up a multi-resolution format for volume data based on tensor approximation, and realized multi-scale feature visualization and multi-resolution volume rendering of 3D data field [138]. Sicat et al. [139] proposed a different representation format of multi-resolution volume data-probability density function (pdfs) sparse volume, which not only allows out-of-core computation of large-scale volume data, however also implements GPU Interactive multiresolution volume rendering. Sicat et al. [139] presented a sparse pdf volume representation for large-scale volume data compactly which accurately encodes the hierarchical pdf information of large multi-resolution volumes as shown in Figure 16, which facilitates consistent multi-resolution volume rendering. In contrast to standard volume compression methods, their method can encode much more information without requiring an impractical amount of additional storage, as well as avoiding explicit decompression for volume rendering. Further algorithm optimization based on multi-resolution also includes detail level selection based on image quality evaluation [140], viewpoint-related data reduction [141], mixed-resolution rendering of highlighted areas [142] and data compression based on wavelet [143].

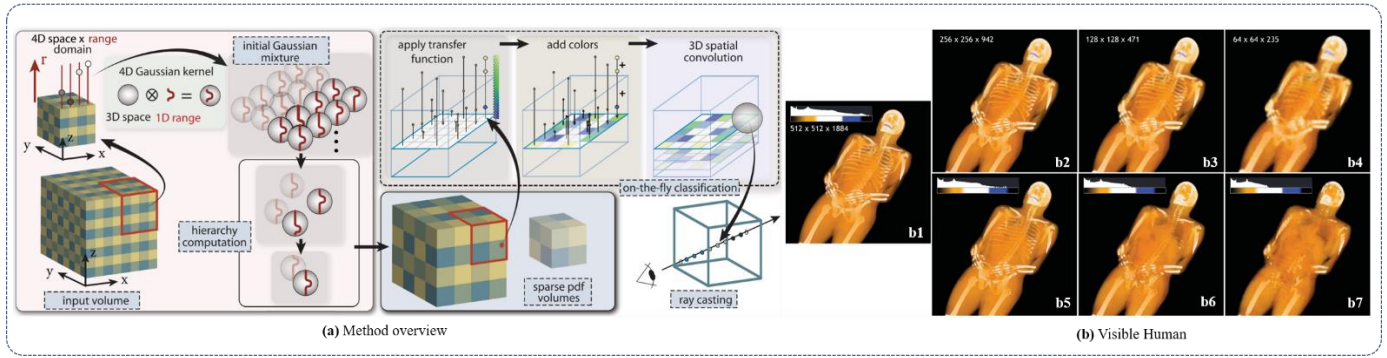

Figure 16: Sparse PDF volumes for consistent multi-resolution volume rendering. (a) Method overview. (b) Visible Human. (b1) Volume rendering level (512×5122 ×1884). Comparison of volume rendering coarser levels (b2, b5)(256×256×942), (b3, b6)(128×128×471), and (b4, b7)(64×64×235) [139].

### 4.2.3 Situ PDF Analysis for Efficient Real-Time Visualization

The close combination and collaboration of numerical simulation and visualization, and the direct data analysis and reduction processing of calculation results based on Probability Distribution Function (PDF) in situ can effectively reduce data transmission and I/O Overhead to meet the needs of timely or real-time visual analysis of large-scale numerical simulation data [144]. Wang et al. [145] of Han-Wei Shen's research group used the probability distribution function based on the Gaussian Mixture model to reduce the data in situ, and then carried out high-quality visualization based on statistics, and proposed to use the Gaussian mixture model (GMM) compresses and stores spatial information, taking into account spatial information not included in previous distribution-based representations. In order to ensure a small storage overhead, an adaptive scheme is used to determine the number of Gaussian components required for each spatial GMM. They qualitatively compared their representations to existing distributed representations. Their methods are able to calculate the probability density function of a value at any location, which represents the possible values and their probability of occurrence, with satisfactory experimental results, with small bias and variance at each voxel. Wei et al. [146] proposed two effective feature search algorithms based on local distribution for multi-field mixed data sets. One is edge feature search, which can provide users with visual exploration of the feature description of each data field. The other is joint feature search, where users can explore joint features based on several attributes in a local area. Dutta et al. [147] proposed a locally isotropic driven block random data summary technology, which can work in situ and retain statistical properties of data through block summary, thus achieving effective probabilistic feature analysis and visualization. Wang et al. [148] proposed a representation of large-scale data analysis based on images and distributions, which can quantify the exploration of transfer functions and uncertainty. They generate proxies from the choices the scientists make about the images and the bandwidth and storage the post-processing machines can handle. With only access to these agents, analysis and visualization can be performed on post-processing machines. Hazarika et al. [149] proposed a flexible distribution-based uncertainty modeling strategy, which was based on a statistically robust multivariate technique called Copula. Using a flexible strategy tailored to the need for uncertainty modeling in scientific datasets, they propose a method for extracting uncertainty/probability features in scalar and vector fields. Figure 17 shows the flowcharts and rendering results of the above three algorithms.

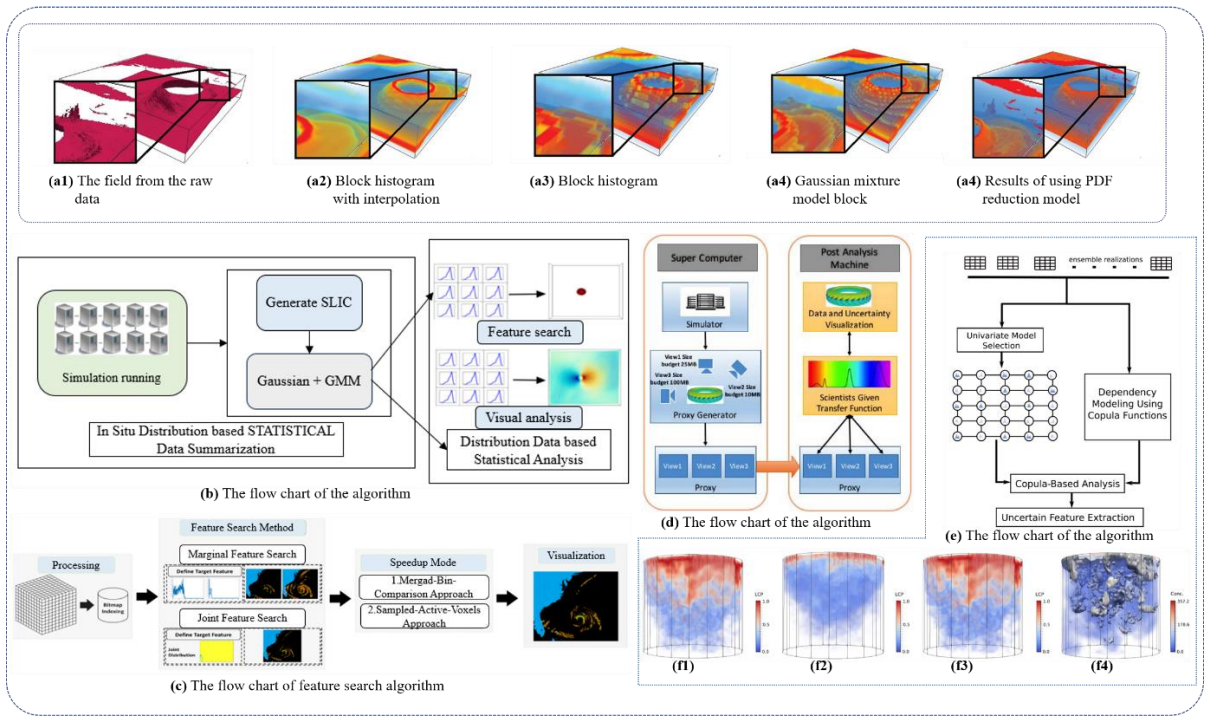

Figure 17. (a) Statistical visualization and analysis of large data using a value-based spatial distribution. (a1)-(a5)Rendering effects using distribution-based representation [148]. (b) Homogeneity guided probabilistic data summaries [147]. (c) Distribution-based feature search in multi-field datasets [146]. (d) Image and distribution based volume rendering for large data sets [148]. (e) Uncertainty visualization using copula-based analysis in mixed distribution models. [149]. (f) Salt Concentration Ensemble: Results of uncertain isocontours representing the viscous fingers for salt concentration level of 50 and generated by (f1) copula-based technique on a mixed distribution field, (f2) assuming Gaussian model at each grid location, (f3) assuming trimodal GMM at each grid location. (f4) shows the shape of an isosurface from a randomly chosen ensemble member [149].

### 4.2.4 Summary

Adaptive rendering, multi-resolution rendering, and PDF reduction are effective methods for improving the efficiency and quality of volume rendering by simplifying the volume data of the target objects.

Adaptive rendering reduces the number of sampling points, thereby decreasing the computational, memory, and communication overheads. It also allows for sampling based on data characteristics, better preserving the key information of the data. However, the design and implementation of current adaptive sampling algorithms are relatively complex. While reducing the number of sampling points, it may be challenging to accurately grasp the variation patterns of physical quantities, leading to potential deviations in the rendered images. Further research can integrate machine learning to develop smarter sampling strategies that better capture the key features of the data. Additionally, optimizing adaptive sampling algorithms in distributed parallel environments can enhance the ability to process large-scale data. Multi-resolution algorithms can significantly improve rendering efficiency when dealing with large-scale data and have good hardware adaptability, enabling efficient volume rendering on hardware with varying performance levels. However, some detail information may be lost at lower resolution levels. To improve the results, image quality assessment techniques can be utilized to automatically select the most appropriate level of detail. Combining AI can lead to smarter multi-resolution rendering strategies. PDF simplification models effectively reduce data transfer and I/O overheads through in-situ data analysis and computation result reduction, maintaining the integrity and accuracy of the data with minimal storage costs. By employing statistical methods, they enhance the quality and accuracy of visualization. However, the adaptability of PDF simplification models is relatively poor, and for certain types of data, specific adjustments and optimizations may be required. In this regard, machine learning can be integrated to achieve smarter PDF simplification and visualization strategies.

### 4.3 Volume Rendering Effect Enhancement

#### 4.3.1 Enhancing Visualization with Depth and Context in Saliency-Driven Volume Rendering

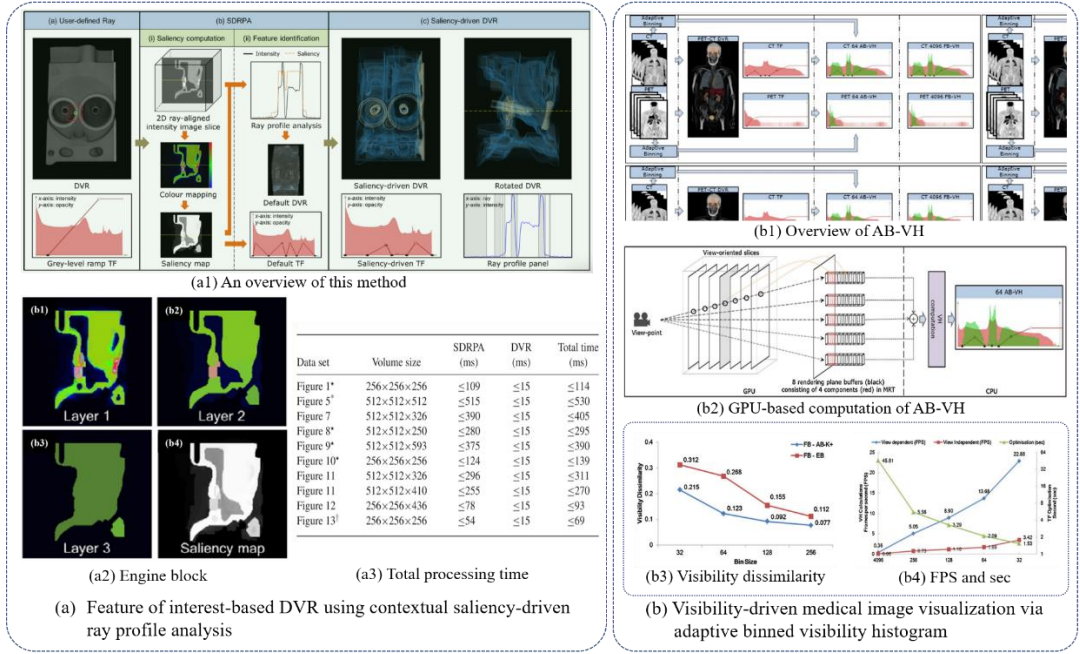

Figure 18: (a) Feature of interest-based DVR using contextual saliency-driven ray profile analysis [152].

The saliency-driven category involves volume rendering of structural segmentation using saliency features such as illumination design, transfer function design, and visualization strategy. [150]proposed a visualization method of "what you see is what you get", which can be simply outlined directly on the volume rendering image, that is, to realize the corresponding modification of the conversion function, and thus designed a variety of rendering effect editing tools. Recently, Zheng et al. [151] established an energy function reflecting the sense of depth order, and adjusted the transparency and illumination parameters by optimizing the energy function, thereby improving the depth order effect in volume rendering and better displaying the internal structure of the data field. Jung, et al. [152] proposed a method for analyzing ray profiles driven by saliency, which utilizes contextual information to identify meaningful features along a user-specified viewing ray. This approach minimizes user interaction during the design of the transfer function and effectively distinguishes structural features that humans deem important. Figure 18 illustrates the principle of the algorithm and the rendering effect. Jadhav et al. [153] introduced Feature Lego, a comprehensive super-voxel clustering framework designed to identify anatomically meaningful regions. The framework first computes super-voxels across the volume and exhaustively clusters these super-voxels to capture all regions and boundaries. It then groups the regions into meta-clusters, allowing users to create desired visualizations by selecting semantic features from the meta-cluster tree. Sharma et al. [154] developed a graph-based approach for designing transfer functions in volume rendering, utilizing graph deduction to identify key structures within the volume. Their method begins by clustering features in the volume data and constructing the topology of the material graph. They then enhance the rendering of important structures by modifying the appearance of user-selected nodes while manually hiding nodes deemed unimportant. Jung et al. [155] introduced an adaptive binned visibility histogram that displays the distribution of visibility across all voxels in a volume. This tool allows users to explore the spatial structures of volume data while receiving real-time feedback about occlusion patterns. It effectively aids users in visualizing specific regions of interest in relation to surrounding structures.

### 4.3.2 Advanced Machine Learning in Predictive Volume Visualization

Researchers have introduced predictive volume visualization techniques to enhance the exploration of internal features in volume data, utilizing advanced machine learning algorithms. Çiçek et al. [156]presented a deep network (3D U-Net architecture) to extract internal features concealed within volume data, learning from sparsely annotated volumetric images. There are two approaches to employing this architecture: the first is a semi-automated method that utilizes user-annotated slices to segment volume data, while the second is a fully automated method that trains on multiple sparsely annotated slices to predict segmentations of volumetric data. The authors have successfully applied their methods to Xenopus kidney data. Jain et al. [157]presented a deep learning method for exploring the internal high-dimensional features within time-varying multivariate volumes. The authors initially train an autoencoder network to learn highly compressed representations, which are subsequently leveraged by the trained decoder network to generate subsampled data. These internal high-dimensional features enable efficient decompression of volume data while preserving high-quality representations. Hong et al. [158] presented a novel volume data visualization pipeline that integrates Generative Adversarial Networks and Convolutional Neural Networks. This approach allows users to interactively explore the implicit internal features of volume data by directly modifying the synthesized images, eliminating the need for complex transfer functions as shown in Figure 19(a). By adjusting visual parameters such as contrast and brightness, users can achieve high-quality volume visualizations. Jönsson et al. [159] presented a cutting-edge visualization system that allows for the interactive exploration of volume data at various levels. Users can navigate through the volume data using several components as shown in Figure 19(b). For instance, the processors generate the bounding box geometry, which enables users to quickly grasp the internal features of volume data.

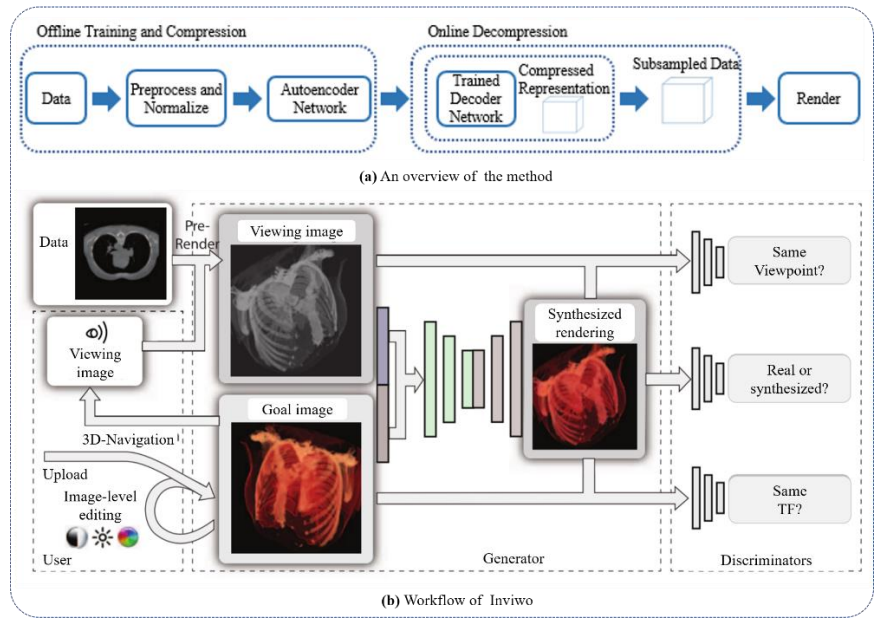

Figure 19: (a) Compressed volume rendering using deep learning [157]. (b) Inviwo—a visualization system with usage abstraction levels [159].

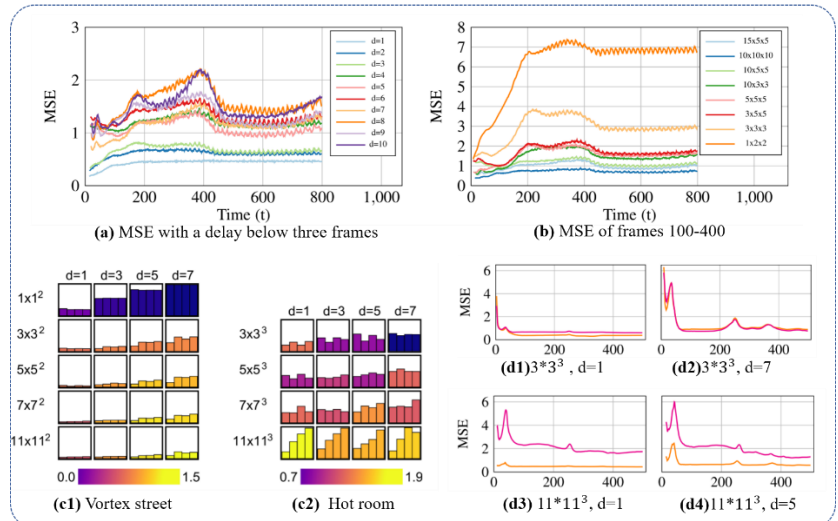

Figure 20: Local prediction models for spatiotemporal volume visualization. [160]

Tkachev et al. [160] developed a novel volume data visualization method that integrates scientific visualization approaches and machine learning approaches to detect and visualize complex internal relationships in spatiotemporal volumes. Figure 20(a-b) shows MSE of different frames and Figure 20(c-d) shows space visualization for two datasets, and temporal misprediction for four different parameter configurations on the hot room dataset. Xu et al. [161] proposed an unsupervised machine learning approach for reconstructing local images in volume rendering. This method focuses on gradient-domain volumetric photon density estimation. It utilizes volume-based features, including albedo, normal, depth, and transmittance, to maintain detailed surface information within the volume data. The resulting renderings achieve high visual quality, ensuring that the details are not obscured by the volume during surface rendering.

### 4.3.3  Models in Realistic Graphics Rendering from Traditional to Deep Learning Approaches

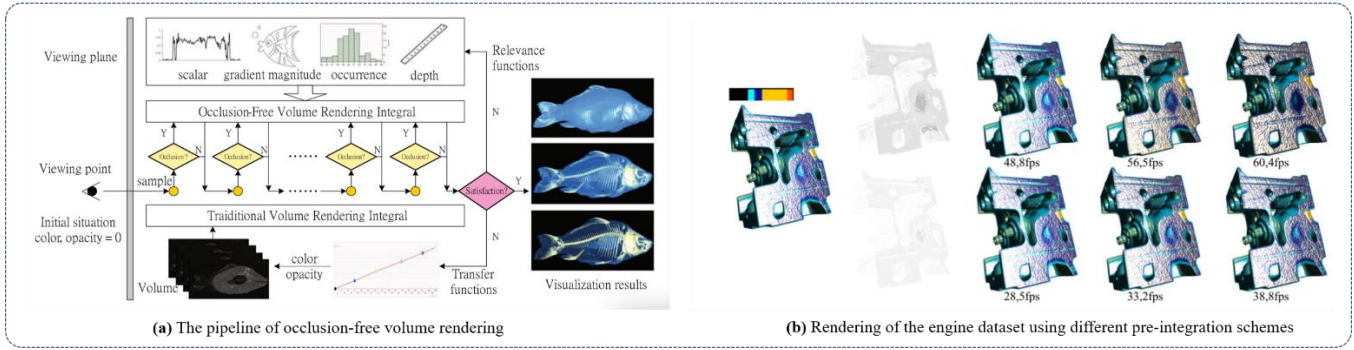

**(a)** The pipeline of occlusion-free volume rendering     **(b)** Rendering of the engine dataset using different pre-integration schemes

Figure 21: (a) Occlusion-free feature exploration for volume visualization [164]. (b) Pre-integrated rendering with non-linear gradient interpolation. [165]

The illumination model serves as the central component in realistic graphics rendering. The most widely used illumination model is the Phong illumination model [162]. Levoy [163] first introduced the Phong illumination model into DVR. The core idea is to use the gradient value of each voxel as the normal vector of the object surface, integrating the gradient into the local illumination model. Zhou et al. [164] introduced a significant function designed to enhance the illumination of a specific surface volume. This function relies on an optimized illumination model that employs a cost function to manage both opacity and light as shown in Figure 21(a). The advanced illumination method improves users' visual perception of key features on the surface. Guetat et al. [165] proposed a nonlinear gradient interpolation method for pre-integral illumination. They assume that the gradient within the pre-integral segment varies along the arc. Based on this assumption, they calculate numerous gradient values within a single pre-integral segment. Additionally, they introduced two-dimensional pre-accumulation tables to enhance the rendering process. Rendering of the engine dataset using different pre-integration schemes is showed in Figure 21(b). Kroes et al. [166] provided a method to reduce ambient occlusion in volume rendering by utilizing a probabilistic approach, which helps avoid the expensive process of ray tracing. As a result, it produces high-quality images and ensures a smooth appearance of volumetric data. With the rapid advancements in machine learning, graphics researchers have also created various machine learning models to automatically generate illumination parameters for specific research tasks. Nalbach et al. [167] presented a convolutional neural network capable of automatically generating screen-space shading effects. This advanced shading technique is entirely determined through deep learning, freeing experts from manually adjusting shading parameters. The outcomes of this deep shading approach provide more detailed insights into the data than traditional methods allow. Engel and Ropinski [168] presented a deep learning method for Volumetric Ambient Occlusion indirect rendering. This approach utilizes a 3D convolutional encoder-decoder architecture to predict volumetric ambient occlusion, complemented by a transfer function that delivers global information. As a result, it offers users a clearer understanding of the overall volume context.

### 4.3.4  Summary

Saliency-driven volume rendering, predictive design volume rendering, and lighting effects are three key techniques in the visualization of volumetric data, optimizing the visualization process through different methods. Saliency-driven volume rendering focuses on identifying and highlighting key features in volumetric data, improving depth-ordering effects and the display of internal structures through lighting design, transfer function design, and visualization strategies. For example, by optimizing energy functions to adjust transparency and lighting parameters, and using contextual information to recognize meaningful features, this method reduces the user's interaction in the transfer function design process and can distinguish important structural features.

Predictive design volume rendering utilizes machine learning to predict and explore internal features within volumetric data. This includes using deep networks to infer internal features, autoencoder networks to learn high-level compressed representations, and combining generative adversarial networks with convolutional neural networks for effective interactive volumetric visualization. These technologies enable users to explore implicit internal features of volumetric data by directly modifying synthetic images, without the complexity of transfer functions. Lighting effects play a central role in rendering, enhancing the illumination effects is an effective to enhance realistic rendering result. For instance, introducing nonlinear gradient interpolation pre-integrated lighting methods, reducing environmental occlusion and using AI to automatically generate lighting parameters could improve the visual perception and understanding of volumetric data.

In future research, the combination of AI and optimization algorithms to automatically adjust saliency and lighting parameters could lead to more natural and accurate visualization of volumetric data. Further automation of the volume rendering process through deep learning could reduce user intervention, increasing efficiency and accuracy. Additionally, applying saliency-driven and predictive design methods to multimodal datasets could provide more comprehensive data analysis and visualization. Integration with CPU parallel computing, GPU, and other specialized hardware acceleration algorithms could also be explored to handle larger datasets and achieve real-time rendering.

## 5. ADVANCED LIGHTING MODELS WITH HIGH PRECISION, REALISM AND EFFICIENCY VOLUME RENDING

Recent experimental studies have demonstrated the feasibility of applying more advanced lighting models to 3D scientific visualization [169, 170, 171]. As research has progressed, over the past few years, interactive volume rendering have begun to support more advanced lighting effects [172, 173]. DVR using volume path tracing (VPT) represents a new trend in volume rendering algorithms that use more advanced physics-based lighting models to produce scientific visualizations that are closer to the real world [174, 175, 176, 177], which is known as cinematic rendering (CR). CR can make the rendering result more realistic. CR [178] introduced a new paradigm for rendering volume data by using physics-based real-time techniques [179]. Traditional volume rendering, such as ray casting [180] , only considers the emission and absorption of energy along the ray to calculate a 3D image, and scattering effects are modeled using a local gradient shadow model, which ignores complex light paths with multiple scattering patterns as well as extinction and lead to less natural and less accurate model. In contrast to traditional VR, CR solves the multidimensional and discontinuous rendering equations by integrating light scattered along the rays from all possible directions [178]. Thus, the path tracking used in CT integrates a large numbers of rays each with a different path to form each pixel of the rendered image. Because the number of optical paths that can be tracked is theoretically infinite, and the tracking of optical paths is computationally expensive, the mainstream CR algorithm uses Monte Carlo simulations to generate a random subset of optical paths with an appropriate distribution. The final image is obtained iteratively by progressively averaging a large number of Monte Carlo samples representing radiosity at random locations and scattering light in random directions [174]. Both VR and CR share the same universal rendering concepts: data segmentation based on voxel decay and the use of color lookup tables that take into account opacity and brightness. Therefore, VR and CR also have the same problem, that is, through the inappropriate use of rendering parameters and adjacent structures may not be able to show the true morphological characteristics of the rendered object [181]. Due to the above similarities between CR and VR, there are no significant differences between the two rendering methods in the visualization of CT image data when used clinically. However, we found that CR has significant improvements in the perception of depth and soft tissue structure, and compared with traditional VR, CR images have more natural and realistic visual effects and better recognition, CR can render more realistic images, and more intuitively let us understand the three-dimensional structure of the rendered model. While CR has potential advantages over VR when it comes to volume data set visualization, CR requires higher computing power due to more complex lighting models. Rendering of the final image takes a few seconds, ranging from 5 to 30 seconds, depending on the quality of the resulting image [182]. Figure 22 shows the difference between MIP, VR and CR.

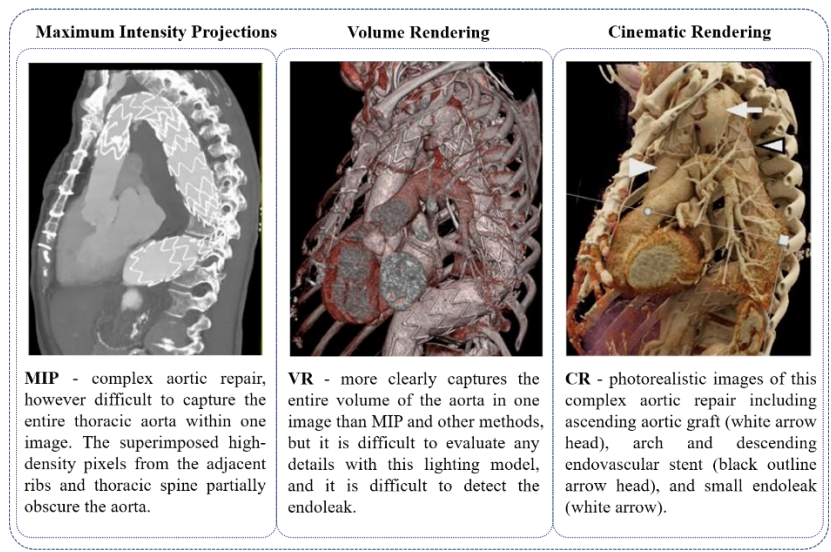

| Maximum Intensity Projections | Volume Rendering | Cinematic Rendering |

**MIP** - complex aortic repair, however difficult to capture the entire thoracic aorta within one image. The superimposed high-density pixels from the adjacent ribs and thoracic spine partially obscure the aorta.

**VR** - more clearly captures the entire volume of the aorta in one image than MIP and other methods, but it is difficult to evaluate any details with this lighting model, and it is difficult to detect endoleak.

**CR** - photorealistic images of this complex aortic repair including ascending aortic graft (white arrow head), arch and descending endovascular stent (black outline arrow head), and small endoleak (white arrow).

Figure 22: The difference between volume rendering and cinematic volume rendering. [183]

CR currently includes various prototype software, the pre-defined transfer functions in the CR package are used to convert the attenuation values of the original data set to the specified color palette and opacity. Therefore, different transfer functions are used to create different levels of clarity [184]. By fine-tuning the transfer function and adjusting the cropping plane, users can highlight different structure and micro-damage. and then capture high quality useful images. although CR is most commonly used in practical applications together with CT, in principle, CR can be performed on any volume imaging data set, while having sufficient ROI Sufficient contrast [183], so CR has considerable prospects in other areas such as industry. Currently, the mainstream CR algorithms are primarily based on Monte Carlo path tracing, and many researchers have also conducted studies on volumetric photon mapping approaches, many-light methods, and other algorithms. They have optimized these algorithms to achieve real-time interactive level CR algorithms and to provide better rendering quality.

## 5.1 Interactive Advanced Quality Rending of Monte Carlo Path Tracing

The global illumination model used in DVR is inspired by the radiative transfer equation, which is the fundamental equation that controls the transmission of light in the participating medium. Kajiya and Von Herzen [185] proposed an approximate solution to this equation for use in computer graphics. Monte Carlo (MC) path tracing is often used to solve this equation in an unbiased manner, with a unified theoretical framework that guarantees convergence to an exact solution. The VPT computes the DVR image by progressively averaging a large numbers of radiation samples evaluated from a randomly selected optical path. The main disadvantage of this algorithm is that producing high quality DVR images requires a lot of rendering time or very expensive hardware equipment to achieve near interactive frame rates. Otherwise, the rendered image exhibits severe noise caused by the MC integration of the sample. The rendering equations [185] used for global illumination can be solved using the path tracing algorithm [186]. A recent investigation by Nov´ak et al. [187] reviewed recent advances in MC path tracing methods for solving optical transmission in participating media. DVR in scientific visualization usually adopts the ray stepping algorithm. For example, Rezk-Salama [188] proposes a GPU-based interactive MC ray casting method for physics-based volume rendering using light travel. Ray travel is simple, however has several disadvantages. For example, it tends to be expensive for high resolution volumes, and rendered images can exhibit unpredictable biases because high-frequency detail can be missed [189, 190]. The researchers accelerated the light-stepping method by using a parallel acceleration method. For example, Shih et al. [191] propose a parallelized, data distributed, and GPU-accelerated algorithm for volume rendering using global illumination. Their approach provides soft shading and rendering on clusters using up to 128 Gpus. In addition, the following methods were used to optimize the efficiency of mc ray-casting. Chaitanya et al. [192] provided an autoencoder architecture for reconstructing global illumination using Monte Carlo methods with extremely low sampling budgets. They show state-of-the-art quality results at interactive rates.

Despite the abundance of research on Monte Carlo path tracing algorithms, current studies have not yet simultaneously addressed the key issues of high computational costs, long rendering times, and severe noise associated with these algorithms. Consequently, some researchers have further optimized and adjusted the traditional Monte Carlo path tracing algorithms to achieve better rendering effects.

### 5.1.1 Data Reduction Requirements Due to Denoising

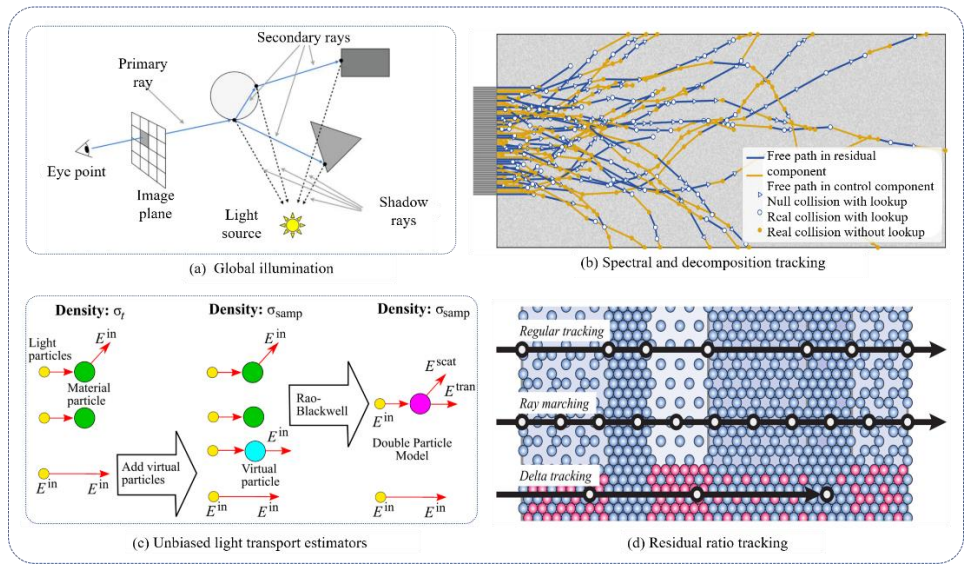

Figure 23: (a) Global illumination principle. (b) Spectral and decomposition tracking for rendering heterogeneous volumes. Kutz's decomposition tracking constructs some of the free paths (yellow) in a heterogeneous medium without evaluating spatially varying collision coefficients. The portion of yellow segments represents the savings in memory lookups. The resulting light paths are identical to delta tracking [196]. (c) Unbiased light transport estimators for inhomogeneous participating media. Adding virtual material particles of Dirac-delta phase function and albedo 1 increases the medium density however preserves the expected radiance. Rao-Blacwellization replaces the three random cases of scattering, transmission and absorption by weighting [197]. (d) Residual ratio tracking for estimating attenuation in participating media. Illustration of different techniques for estimating transmittance through slabs of different materials. Regular tracking finds intersections with interfaces between individual materials. Ray marching proceeds with a constant step that needs to be small enough to avoid excessive bias. Delta tracking fills optically thinner regions with fictitious particles (red), analytically samples tentative free paths, and probabilistically decides whether collisions occur with real or fictitious particles [190].

An alternative to ray-stepping is incremental tracking, a type of critical sampling that determines the free path based on the probability density function (PDF) corresponding to the optical depth in the participating medium. Woodcock tracking [193] is a widely adopted unbiased solution that adjusts the sampling distance to be small enough to properly sample dense areas in the volume. This algorithm has been re-examined in offline rendering for adaptive sampling on large-scale sparse non-uniform media [194], and optimized for CR [195], free path sampling with probability not necessarily proportional to volume transmission has been achieved using weighted incremental tracking methods [196, 190, 197], Figure 23(b-d) shows the principle of these methods. All these methods help reduce noise in the estimated optical path of the participating medium. However, in applications requiring real-time rendering, the aforementioned sampling strategies may require a large number of samples, and free path sampling might not meet the requirements for real-time performance.

### 5.1.2 Progressive MC Path Tracing for Noise-free

Kroes et al. [179] and Liu et al. [198] demonstrated that progressive VPT using GPUs can achieve interactive frame rates for unbiased rendering. Figure 24 shows the flowchart of progressive MC path tracking. Although progressive VPT can converge to noise-free images, it produces very noisy results for interactively rendered scenes and requires expensive distributed rendering systems when rendering [176, 191].

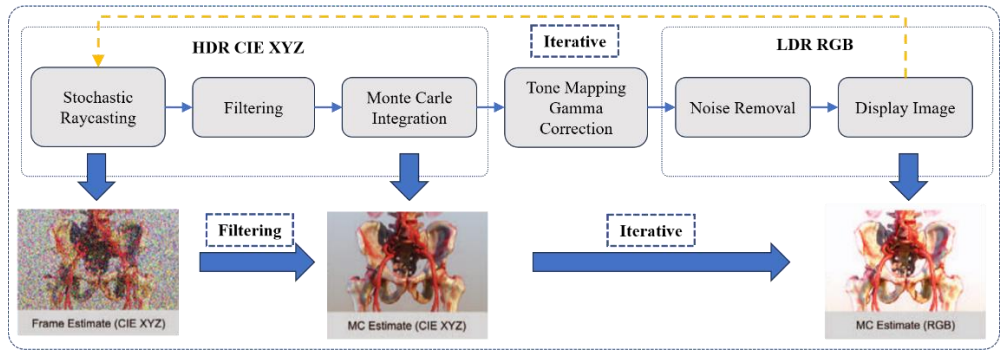

Figure 24. A overview of progressive MC path tracking. [179]

### 5.1.3 Real-Time De-noising for MC Path Tracing

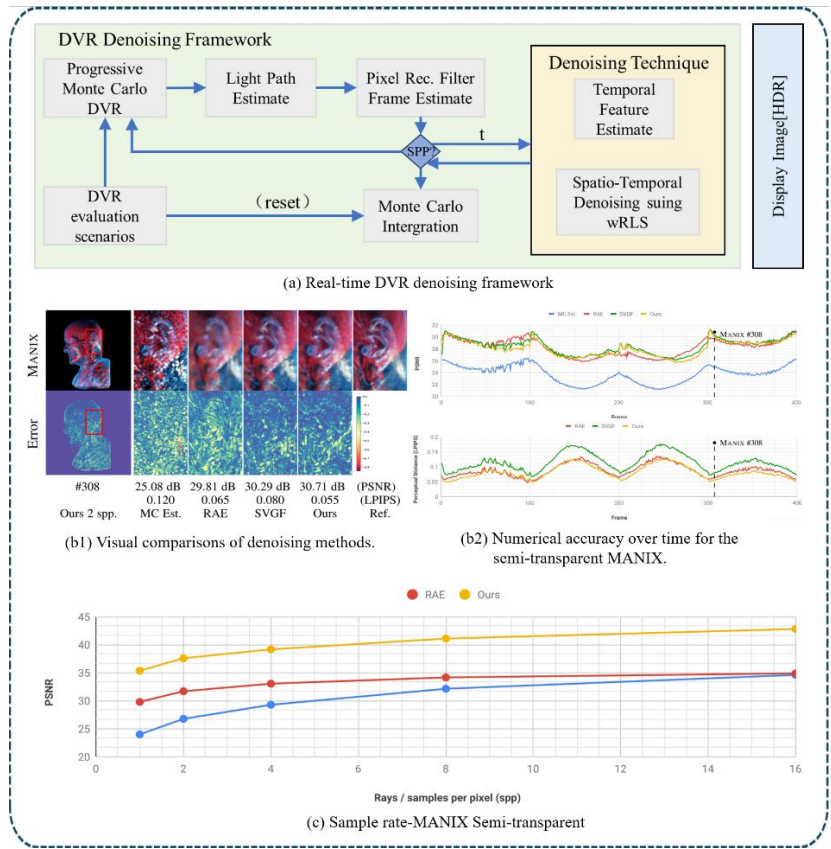

Figure 25: Real-time denoising of volumetric path tracing for direct volume rendering. [206]

Image-space reconstruction has been widely recognized as a feasible method to reduce MC path tracking noise in surface models. Zwicker et al. [199] conducted a comprehensive survey of this topic. Recently, Schied et al. [200] proposed a real-time spatiotemporal denoising technique, which can accumulate pixel colors of each frame and use color variance to control smoothness. Mara et al. [201] designed a real-time denoising device specifically designed to reduce noise on matte and bright surfaces. These techniques are specifically used to filter the global light noise of surface models. De-noising of volume and participating media is often associated with producing offline rendered scenes [202]. Deep learning denoiser have been gaining popularity recently [203, 204]. For example, Chaitanya et al. [192] proposed a deep learning interactive denoiser for MC path tracking. However, these denoisers for surface models rely heavily on noise-free G-buffers (which are generally not available for volume) to produce high-quality denoised images. In the context of DVR, Kroes et al. [179] used a general noise reduction filter in their GPU implementation, however this general filter did not effectively remove MC variation and time flicker. Due to the presence of noise in the G-buffer information used to reconstruct image details, the application of specialized surface model denoising in VPT (such as RLS adaptive denoising [205] ) may not be effective. Therefore, Jose A et al. [206] designed a real-time de-noising method as shown in Figure 25(a) for volume path tracking for DVR to reduce variance noise and time flicker of VPT interactive DVR without relying on problematic G-buffers. This technique does not require any prior training to produce time-stable results for different types of user interactions. Figure 25(b1), (b2) and c, Jose A's result achieves good reconstructions in comparison to RAE and SVGF, however much better temporal stability and numerical convergence. While the aforementioned algorithms can reduce noise to some extent, they still fail to address the high computational cost of Monte Carlo path tracing algorithms, and may even lead to increased rendering times.

## 5.2 Realistic Rendering of Volumetric Photon Mapping

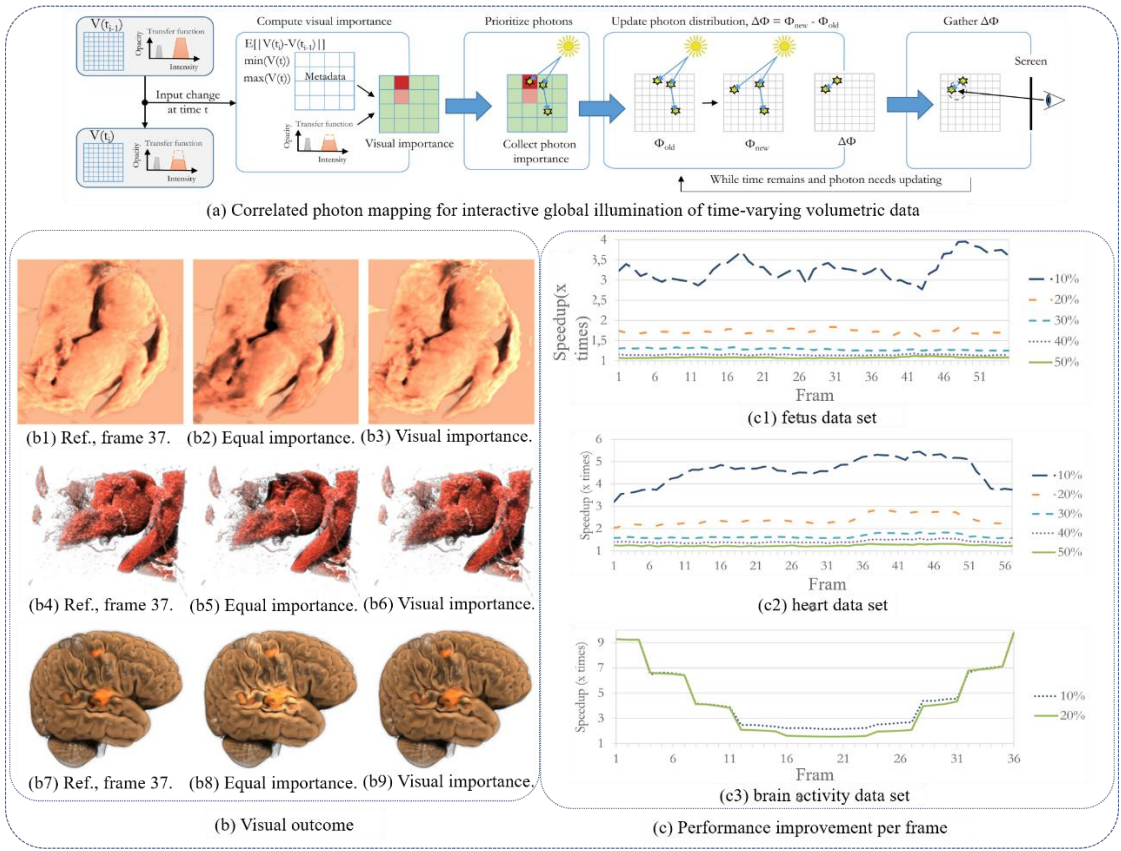

(a) Correlated photon mapping for interactive global illumination of time-varying volumetric data

(b1) Ref., frame 37. (b2) Equal importance. (b3) Visual importance.

(b4) Ref., frame 37. (b5) Equal importance. (b6) Visual importance.

(b7) Ref., frame 37. (b8) Equal importance. (b9) Visual importance.

(b) Visual outcome

(c1) fetus data set

(c2) heart data set

(c3) brain activity data set

(c) Performance improvement per frame

Figure 26: Correlated photon mapping for interactive global illumination of time-varying volumetric data. [213]

Volume photon mapping [207] and progressive extension [208] share expensive calculations to solve volume rendering integrals by caching optical transfers. Jarosz et al. [209] J introduced a variant of Woodcock tracking in the progressive photon beam, which is an extension of the photon beam in volume photon mapping [210]. Jönsson et al.[211] implemented interactive DVR through photon mapping, which recalculated only the changed photons. However, their photon collection phase is computationally expensive, resulting in a lower frame rate when the camera is moved. Zhang et al. [212] proposed a predictive volume irradiance transmission method for predictive photon mapping using basis function coding. Although this method enables real-time radiation reconstruction, the photon map is regenerated every time the transfer function changes. In order to speed up the generation of the photon map, Jönsson and Ynnerman [213] provided an interactive global illumination method that leverages the correlation of photon maps during the changes of the volumetric data or visual parameters. As shown in in Figure 26(a), the authors use visually de-correlated areas in the volume to effectively update the photon, while they use the transfer function to determine the important regions of the volume data. Figure 26(b) and (c) proved this method can perform high-quality rendering results of the significant important region of a volume surface, while maintains visual fidelity, as well as performance.

Volumetric photon mapping simulates the propagation of light through volumetric media by emitting photons and tracking their interactions with the scene, achieving volume rendering. This method is particularly suitable for simulating phenomena such as scattering and caustics of light in volumetric media, which are difficult to handle efficiently with traditional lighting methods. Compared to traditional ray tracing, the artifacts produced by volumetric photon mapping are typically low-frequency and less noticeable, thereby enhancing the realism of the rendering. However, volumetric photon mapping requires the storage of a large amount of photon data and has high computational complexity, which can lead to significant memory consumption and computational costs. Although volumetric photon mapping offers excellent rendering effects, its efficiency is relatively low. Therefore, the current main research direction is to improve rendering speed through algorithm optimization, adaptive techniques, and machine learning technologies.

### 5.3 High Realism and flexibility of Many-light methods

Like photon mapping, the multi-light method is a bi-directional MC technique. Photon mapping relies on density estimation and needs to track a large number of photons, while multi-light methods require orders of magnitude fewer light paths, so rendering efficiency is very high. Engelhardt et al. [214] describe a particle tracking algorithm for creating a set of virtual point sources (VPL) in the participating media and derive a GPU-friendly bias compensation scheme for high-quality rendering. Zhang [215] effectively improved the rendering effect by using three-point lighting and background light. Yuan et al. [216] designed a volume rendering algorithm that supports multiple light sources and can simulate rapid propagation, absorption and scattering of multiple light sources at the same time. The system injects a ray of light in each of the six directions of the sampled volume data. As shown in the figure 27, the rendering effect of 6 light sources is significantly improved than that of 2 light sources and 4 light sources. Weber et al. [217] applied a multi-light source approach using VPL to interactive volume rendering. In particular, the technology is tailored for interactive editing of volume transfer functions, providing instant updates and redistributing contributions from VPL. However, interactive visualizations limit the number of VPLS, and transfer function edits are limited to smooth transitions, as they require further reallocation and recalculation of VPLS. Time consistency is improved by gradually updating the location of the VPL and refreshing only its incremental contributions. However, major changes (such as switching to a completely different transfer function) can cause noticeable flickering.

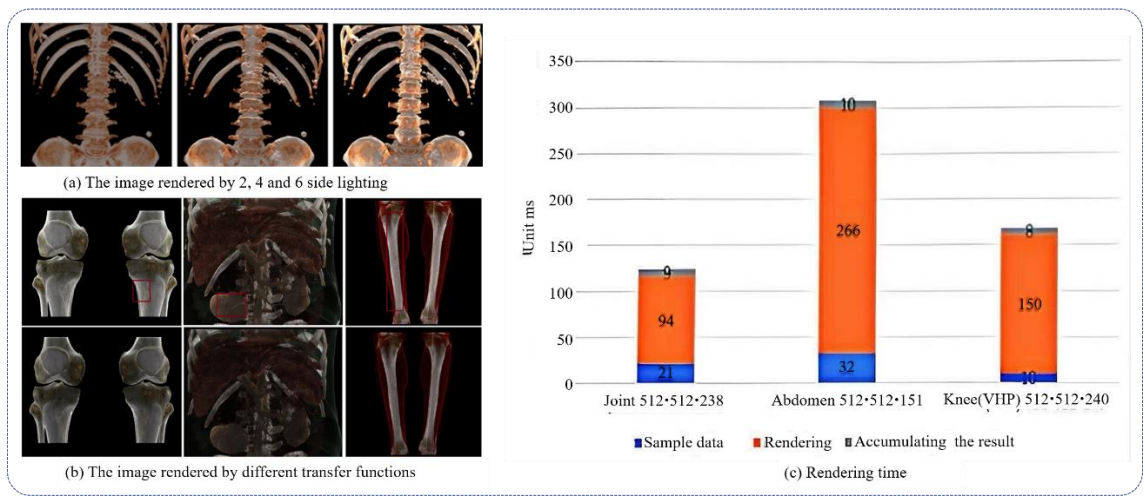

(a) The image rendered by 2, 4 and 6 side lighting

(b) The image rendered by different transfer functions

(c) Rendering time

Figure 27: Cinematic volume rendering algorithm based on multiple lights photon mapping. [216]

The multi-light method enhances the realism of a scene by simulating the illumination effects of multiple light sources on the scene, thus offering high flexibility. By controlling the intensity, color, and direction of different light sources, a variety of lighting effects can be created, such as shadows, highlights, and reflections. Additionally, because different light sources can produce shadows from various directions, they provide better depth information, helping observers better understand the three-dimensional structure of the scene. However, involving multiple light sources increases the computational cost of rendering, as each light source may affect every pixel in the scene, leading to increased rendering times. Moreover, managing and adjusting multiple light sources can become complex, especially when there are many light sources or the scene is complex. This can lead to rendering artifacts such as light spots, overexposure, or unnatural shadows, which may affect the identification and judgment of minor damages. Therefore, to meet the needs of real-time interaction, it is necessary to research and develop new algorithms and technologies to achieve real-time or near-real-time rendering under multi-light conditions. GPU and other specialized hardware can be utilized to accelerate multi-light rendering and improve efficiency. At the same time, artificial intelligence and machine learning technologies can be used to automatically optimize light source parameters, reducing the need for manual adjustments to achieve higher efficiency and to minimize unnatural rendering results.

### 5.4 Precise, Realistic and Efficient Diffusion Approximations

For rendering multi-scattering global illumination effects in participating media, the method based on diffusion approximation [218] is an effective alternative to MC path tracking. K¨orner et al. [219] proposed the Flux-Limited Diffusion (FLD) technique, which improved the Classical Diffusion Approximations (CDA) for heterogeneous media. CDA The method is affected by non-physical radiation flux in the transparent region, and FLD can produce more accurate results than CDA when compared with path tracking ground reality as shown in Figure 28. Although the proposed FLD solvers converge faster than MC path tracing or photon mapping, progressive or interactive scaling has not been proposed.

Diffusion approximations simulate complex physical processes through approximate calculations to accelerate rendering speeds. This method can better simulate the scattering and attenuation of light, making the rendered images more realistic. It also adapts to different datasets and rendering conditions, offering a certain degree of flexibility. However, approximate calculations may lose details in some cases, leading to rendering results that deviate from the actual physical processes. Additionally, diffusion approximation models may require meticulous parameter adjustments to achieve the best results, which can increase the difficulty of use. Currently, research on diffusion approximations in the field of volume rendering is not mature, and there is a lack of in-depth studies.

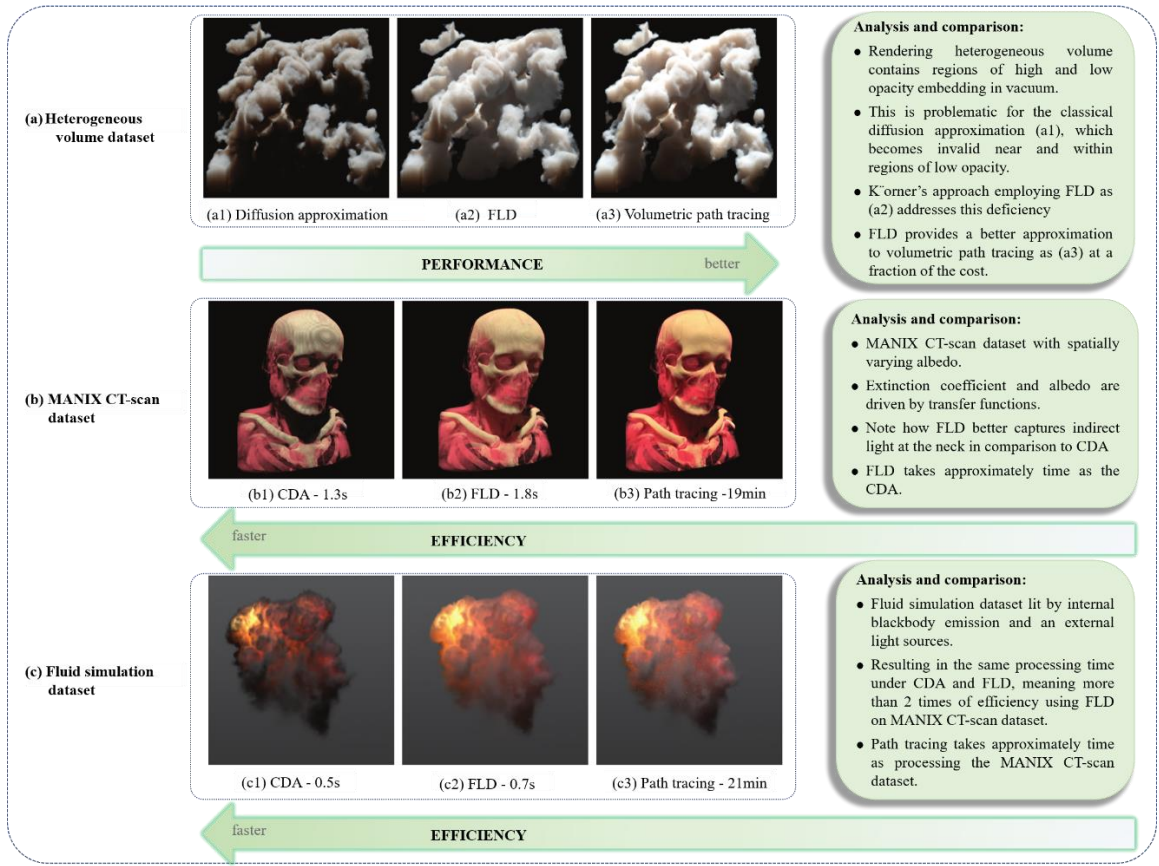

Figure 28: Flux-limited diffusion for multiple scattering in participating media. [219]

### 5.5 High Precision, Realism and Efficiency of Irradiance Caching

The irradiance cache uses smoothly varying indirect illuminance to pre-calculate irradiance transfer within the volume. In an earlier study, Kajiya and Von Herzen [185] proposed a two-channel approach for simulating the global illumination effects of heterogeneous datasets. In the first step, the radiosity of each voxel is estimated and continuously integrated along the view rays in the second step. However, the first step takes a long time and therefore is not suitable for interactive visualization. Alternatively, the estimated irradiance can be calculated only at sparse cache points in the volume. For example, Kriv´anek et al. [220, 221] proposed the use of spherical harmonics (SH) to store and interpolate orientation-dependent irradiance. Later, Jarosz et al. [222] extended this approach to participating media. Kronander et al. [223] obtained real-time performance of DVR by encoding local and global volume visibility using SH on a multi-resolution grid. Recently, Khlebnikov et al. [224] proposed parallel irradiance caching and MC path tracking for interactive volume rendering. However, the irradiance cache stores and updates pre-calculated irradiance, which affects certain interactivity. Jones and Reinhart [225]proposed a novel method of parallel irradiance caching for global illumination on a graphics processing unit (GPU), which can generate images similar to those created by Radiance's RPICT program up to twenty times faster and have better effect.

Irradiance caching enhances rendering speed by reusing lighting information within a scene, reducing the need for lighting calculations for each pixel. It also produces smooth lighting effects because it estimates lighting through interpolation and filtering techniques, which reduces noise and discontinuities in the rendering process. Irradiance caching is particularly effective for scenes where indirect lighting changes slowly, as it can capture and reuse the low-frequency components of lighting. However, irradiance caching is viewpoint-dependent, meaning that when the camera or scene changes, the cached data may no longer be valid and need to be recalculated. This conflicts significantly with the current needs for real-time interactivity, making it a primary issue that future research should address. Figure 29(a) and Figure 29(b) show the performance of the rendering algorithm in enclosed spaces and open spaces.

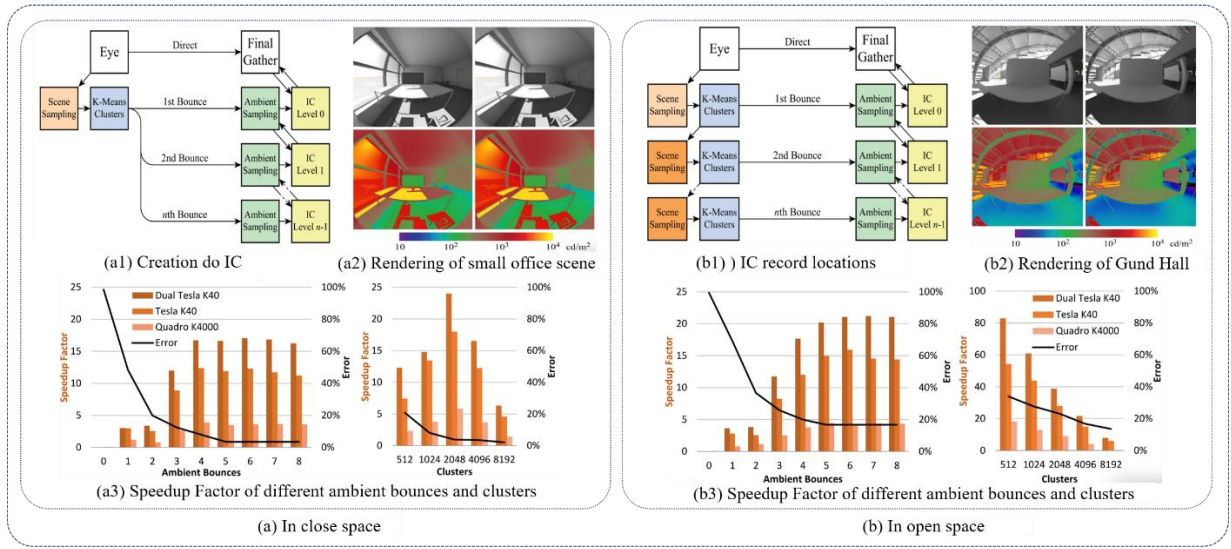

Figure 29: Irradiance caching for global illumination calculation on graphics hardware. (a) In enclosed spaces. (a1) IC record locations from a single call to the sampling kernel are used to create the IC at each level. (a2) The small office scene rendered with five ambient bounces in RPICT (left) and 17 times faster in OptiX™ implementation (right). (a3) For the small office scene, the speedup facto increases and error decreases with the number of ambient bounces using 4096 clusters (left). Error decreases with the number of clusters, however large numbers of clusters require greater traversal time using five ambient bounces (right). (b) In open spaces. (b1) IC record locations are separately calculated for each IC level based on locations reached at the previous level. (b2) The Gund Hall scene rendered with five ambient bounces in RPICT (left) and 20 times faster in OptiX™ implementation (right). (b3) For Gund Hall, the speedup factor increases and error decreases with the number of ambient bounces using 4096 clusters (left). Error decreases with the number of clusters, however large numbers of clusters require greater traversal time using five ambient bounces (right). [225]

In summary, recent studies have demonstrated the potential of incorporating advanced lighting models into 3D scientific visualization and interactive volume rendering. DVR and VPT, along with CR, represent new trends that utilize more sophisticated physics-based lighting models. These methods aim to produce scientific visualizations that are closer to real-world appearances. Traditional volume rendering techniques, such as ray casting, are limited in their ability to calculate the emission and absorption of energy along rays, often leading to less natural and less accurate virtual reality (VR) images. In contrast, CR approaches the multidimensional and discontinuous rendering equations by integrating light scattered along rays from all possible directions, resulting in more realistic images but at the cost of higher computational requirements and longer rendering times. The rendering of the final image can take from several seconds to half a minute, depending on the desired quality. Research is ongoing to optimize these algorithms for real-time interactive level CR algorithms and to enhance rendering quality. This includes exploring volumetric photon mapping, many-light methods, and other algorithms to achieve real-time or near-real-time rendering under multi-light conditions. Techniques such as diffusion approximations and irradiance caching are also being investigated to simulate global illumination effects and to increase rendering speeds, respectively.

However, these advanced rendering techniques face challenges such as high computational costs, noise, and the need for interactivity, which are driving further research and development in the field of volume rendering. The goal is to balance realism, precision, and efficiency while addressing the limitations of current methods to meet the demands of real-time interactivity in various applications, including medical imaging and scientific visualization [226-230].

## 6. CONCLUSION

This review has extensively deliberated on the emergence and progression of 3D visualization predicated on Micro-CT for volume datasets. The confluence of CT reconstruction and volume rendering is pivotal in determining the fidelity and expeditiousness of 3D visualization. Furthermore, the review has introduced an avant-garde volume rendering algorithm that incorporates global illumination and path tracing. The discussions encompass the following key points:

(1) Micro-CT, a form of radiographic imaging, discerns the morphological structure of target materials without causing specimen damage, aligning well with the nuances of volume rendering to better reveal the internal intricacies of objects. The light sources for Micro-CT are primarily categorized into synchrotron radiation and micro-focus X-ray source. The former, when utilized with a Fresnel zone plate, can achieve a resolution of 10nm; however, the scarcity of synchrotron radiation CT systems globally (numbering only in the dozens) has rendered the more accessible laboratory Micro CT system as the mainstream option. With a minimum resolution of 0.5 microns, it has been extensively implemented in the study of microstructures. Nevertheless, the pursuit of higher resolution, while enhancing imaging quality, also engenders larger datasets and diminished reconstruction efficiency, posing a significant challenge for the realization of digital twins.

(2) This review summarizes the classification and development of major weight reconstruction algorithms, and provides an overview of the current effects and efficiency of reconstruction algorithms. FBP is known for its fast reconstruction speed and good image texture, but it has larger noise and artifacts, and it requires a relatively complete set of original data, which also means a longer scanning time, posing a significant obstacle in the implementation of digital twins. Hybrid-IR and MBIR also belong to IR and have similar characteristics, performing better than FBP in noise reduction, low-contrast object resolution, and dose and artifact reduction, but with slower reconstruction speeds and slightly worse image texture due to excessive smoothing. DLR, which combines the above algorithms with AI, shows excellent performance in noise reduction, low-contrast object resolution, scanning efficiency, artifacts, and maintaining spatial resolution. It demonstrates strong potential in clinical medicine, providing fast reconstruction speeds, better dose performance, and less noise and artifacts, and also has great potential in the industrial field. However, the issue of ghost artifacts that may arise during the DLR reconstruction process still needs to be addressed, as they could affect the judgment of defects. At the same time, DLR also faces the problem of a lack of industrial training datasets.

(3) This review summarizes the current state of research on two key issues in large-scale data volume rendering: rendering efficiency and rendering effect. With the development of scientific computing and the increase in HPC computing power, the amount of data generated by scientific and engineering simulations has been growing larger and larger. Traditional scientific visualization methods are increasingly struggling to handle and analyze data from petaflop-scale computations. To address the performance bottleneck in the analysis of ultra-large-scale computational data, researchers have proposed three types of optimization algorithms based on ray casting: volume rendering acceleration, volume rendering data simplification, and volume rendering effect enhancement. The first two types of algorithms mainly improve the efficiency of volume rendering, while the third type focuses on the quality of volume rendering. Volume rendering acceleration make better use of computer hardware, improving rendering processing efficiency through parallel rendering and GPU hardware acceleration. Depending on the processing object, different algorithms can achieve a 6-14 times increase in efficiency. Volume rendering data simplification process the target data volume while ensuring rendering quality through adaptive rendering, multi-resolution rendering, and PDF simplification models, reducing the overhead of computation, transmission, and communication, and improving rendering efficiency to meet the needs of real-time visualization analysis for large-scale datasets. To achieve high-fidelity rendering effects, volume rendering effect enhancement conduct research and optimization from feature extraction and effect enhancement, achieving good results in rendering quality. Better rendering effects can provide domain experts with more information to draw more accurate conclusions. The aforementioned methods have significantly improved both the quality and efficiency of volume rendering.

(4) In order to further enhance the realism and effectiveness of volume rendering, researchers have proposed a volume rendering algorithm based on path tracing, which utilizes more advanced physical lighting models to generate scientific visualization effects that closely resemble the real world. This algorithm, known as CR in medical imaging, significantly improves the perception of depth compared to traditional volume rendering, resulting in images that are more natural, realistic, and have higher recognizability. However, CR requires greater computational power, and the rendering of the final image may take several seconds to tens of seconds. Moreover, the main CR algorithms are based on Monte Carlo path tracing, which can lead to high computational costs, long rendering times, and severe noise issues. Researchers have optimized and adjusted CR to achieve better rendering effects through improved sampling strategies, progressive Monte Carlo path tracing, and image denoising. Additionally, this review also covers volume photon mapping methods, multi-light source methods, and diffusion approximations in CR algorithms. Although these techniques have made significant progress in rendering quality, they still face challenges such as high computational costs and long rendering times. Future research will need to address these primary issues related to real-time interactivity requirements for structural health monitoring (SHM).

## 7. PROSPECT

While there remains a discernible gap between current CT equipment and 3D visualization algorithms and the realization of real-time digital twins, 3D visualization based on CT continues to be a principal modality for non-destructive internal inspection of objects. The entire 3D visualization process encompasses multiple steps, and with advancements in science and technology, addressing the core issues at each step could potentially realize real-time digital twins.

(1) The pixel size of the sensor in micro-CT determines the upper limit of the CT equipment's resolution and also the upper limit of the quality of three-dimensional imaging. This is related to the advanced manufacturing processes of semiconductors. Currently, top semiconductor processes (such as TSMC's 3nm process) are mainly used in mobile chips or high-computing power chips due to cost considerations. As semiconductor manufacturing processes improve and become more widespread, Micro-CT can detect more microscopic damage or physical characteristics. At the same time, future Micro-CT can integrate multimodal data, combining optical microscopes, electron microscopes, and other imaging techniques to achieve multi-scale and multi-dimensional information integration, obtaining higher information density and providing more comprehensive sample analysis.

(2) DLR has undoubtedly emerged as the primary research focus among current reconstruction algorithms, However，DLR has certain limitations：relying heavily on high-quality, large-scale training sets, and the trained singular algorithm lacks generalization capabilities, failing to be universally applicable. Future research may concentrate on enhancing the algorithm's generalization and robustness, employing heterogeneous datasets for training to ensure stable operation across diverse environments and conditions. Furthermore, as AI continues to evolve, DLR algorithms may achieve dynamic learning, optimizing their performance based on new data post-installation, rather than static learning during the training phase.

(3) Large-scale datasets often demand substantial computing resources. However, in practical applications, scientific computing tasks occupy the majority of computational resources, leaving only about one-tenth of the total resources for visualization tasks. This allocation does not make full use of the hardware resources. To better adapt to the complex architecture of high-performance computers, hybrid parallel rendering that involve inter-node multi-process data parallel processing and intra-node multi-core coupled accelerator multi-threaded local data parallel processing can be developed. These techniques can be used to accelerate volume rendering algorithms at multiple levels, thereby improving the efficiency of volume rendering.Currently, parallel volume rendering algorithms mainly apply uniform sampling or stratified uniform sampling, without effectively setting sampling points based on the intrinsic characteristics of the data, thus wasting computational and storage resources. To conserve valuable computational resources, researchers can study feature analysis and extraction techniques tailored to the complex data features in specific applications, focusing on presenting features of interest to users and developing enhancement for complex feature volume rendering to improve the sense of hierarchy and realism of feature structures. Besides, volume data can be simplified by deeply mining the intrinsic characteristics of the data, reducing the amount of data beyond key features, accurately grasping the fine physical features in the data field, and further developing adaptive volume rendering algorithms to significantly enhance the rendering efficiency of data features. Additionally, volume rendering algorithms can be combined with machine learning and optimization algorithms to automatically adjust significance and lighting parameters, achieving a more natural and accurate visualization of volume data. By leveraging deep learning to further automate the volume rendering process, thereby improving efficiency and accuracy.

(4) CR, which demands substantial computing resources, has yet to become the preferred choice for interactive volume rendering. Compared to traditional volume rendering, CR can provide visualization results closer to real-world scenarios. With future advancements in computing power, CR may be integrated with VR and AR to achieve a comprehensive and multi-dimensional visualization effect. Concurrently, as network technology evolves, researchers may remotely access and analyze 3D models, fostering remote communication and collaboration among experts across various geographical locations to enhance research efficiency.

## ACKNOWLEDGMENTS

This work was supported by the National Nature Science Foundation of China (No. 62171271).

**Declaration of interests**

☒The authors declare that they have no known competing financial interests or personal relationships that could have appeared to influence the work reported in this paper.

☐The authors declare the following financial interests/personal relationships which may be considered as potential competing interests: